\documentclass{hld2024} %

\PassOptionsToPackage{compress,square}{natbib}

\usepackage[utf8]{inputenc} %
\usepackage[T1]{fontenc}    %
\usepackage{hyperref}       %
\usepackage{booktabs}       %
\usepackage{amsfonts}       %
\usepackage{nicefrac}       %
\usepackage{microtype}      %
\usepackage{xcolor}         %
\usepackage{color}\definecolor{darkblue}{rgb}{0,0,0.75}
\usepackage{url}
\usepackage{xcolor}
\usepackage{listings}

\usepackage[textsize=tiny,textwidth=2.5cm]{todonotes}

\usepackage{subfiles}

\title[Landscaping Linear Mode Connectivity]{\vspace{-3mm}Landscaping Linear Mode Connectivity}

\hldauthor{%
\Name{Sidak Pal Singh}, \addr {ETH Zürich, Switzerland} 
\Email{ssidak@ethz.ch}
\AND
\Name{Linara Adilova},
\addr {Ruhr-Universität Bochum, Germany} 
\Email{adylova.linara.r@gmail.com}
\AND
\Name{Michael Kamp},
\addr {Institute for AI in Medicine IKIM, Germany} 
\Email{info@michaelkamp.org}
\AND
\Name{Asja Fischer}, \addr {Ruhr-Universität Bochum, Germany}
\Email{asja.fischer@rub.de}
\AND
\Name{Bernhard Schölkopf}, \addr {MPI-IS Tübingen, Germany} 
\Email{bs@tuebingen.mpg.de}
\AND
\Name{Thomas Hofmann}, \addr {ETH Zürich, Switzerland} 
\Email{thomas.hofmann@inf.ethz.ch}
}

\usepackage{float}
\usepackage{amsmath}
\usepackage{dsfont}
\usepackage{amssymb}
\usepackage{bm}
\usepackage{nicematrix}
\usepackage{hyperref}
\hypersetup{
	colorlinks=true,
	linkcolor=blue,
	filecolor=magenta,      
	urlcolor=magenta,
	citecolor=blue,
}

\usepackage{mdframed}
\usepackage{enumitem}

\usepackage[labelfont=bf]{caption}
\usepackage{booktabs}

\newtheorem{proposition1}[theorem]{Proposition}

\newtheorem{assump}{Assumption}

\makeatletter
\newtheorem*{rep@theorem}{\rep@title}
\newcommand{\newreptheorem}[2]{%
\newenvironment{rep#1}[1]{%
 \def\rep@title{#2 \ref{##1}}%
 \begin{rep@theorem}}%
 {\end{rep@theorem}}}
\makeatother

\newreptheorem{proposition1}{Proposition}
\newreptheorem{lemma1}{Lemma}
\newreptheorem{theorem1}{Theorem}
\newreptheorem{corollary1}{Corollary}
\usepackage{bbm}

\def\eqref#1{eq.~\ref{#1}}

\def\1{\bm{1}}

\newcommand{\btheta}{{\pmb \theta}}

\newcommand{\Loss}{{\mathcal{L}}}

\usepackage{mdframed}

\usepackage{mdframed}

\usepackage{thmtools}
\usepackage{cleveref}

\declaretheoremstyle[%
spaceabove=-2pt,%
spacebelow=6pt,%
headfont=\normalfont\itshape,%
postheadspace=1em,%
qed=\qedsymbol%
]{mystyle}

\newcommand*{\colorboxed}{}
\def\colorboxed#1#{%
	\colorboxedAux{#1}%
}
\newcommand*{\colorboxedAux}[3]{%
	\begingroup
	\colorlet{cb@saved}{.}%
	\color#1{#2}%
	\boxed{%
		\color{cb@saved}%
		#3%
	}%
	\endgroup
}

\newreptheorem{proposition}{Proposition}
\newreptheorem{lemma}{Lemma}
\newreptheorem{theorem}{Theorem}
\newreptheorem{corollary}{Corollary}
\newreptheorem{assump}{Assumption}

\makeatletter
\newcommand{\neutralize}[1]{\expandafter\let\csname c@#1\endcsname\count@}
\makeatother

\usepackage{graphicx}
\usepackage{wrapfig}
\usepackage{mathtools} %
\usepackage{sidecap}
\usepackage{subcaption}

\usepackage[most]{tcolorbox}
\tcbset{boxsep=0mm,boxrule=0pt,colframe=white,arc=0mm,left=0.5mm,right=0.5mm}

\usepackage{soul}
\usepackage[normalem]{ulem}

\newcommand{\Pm}{{\mathbf{P}}}
\newcommand{\soln}[1]{{\btheta^\ast_{#1}}}
\newcommand{\barrier}[1]{{\mathcal{B}({#1}})}

\usepackage{dsfont}
\newcommand\dsone{\mathds{1}}
\newcommand*{\QEDA}{\hfill\ensuremath{\blacksquare}}

\begin{document}
\vspace{-5mm}
\maketitle
\vspace{-8mm}
\begin{abstract}%
The presence of linear paths in parameter space between two different network solutions in certain cases, i.e., linear mode connectivity (LMC)~\citep{frankle2020linear}, has garnered interest from both theoretical and practical fronts. There has been significant research that either practically designs algorithms catered for connecting networks by adjusting for the permutation symmetries as well as some others that more theoretically construct paths through which networks can be connected~\citep{kuditipudi2020explaining}. Yet, the core reasons for the occurrence of LMC, when in fact it does occur, in the highly non-convex loss landscapes of neural networks are far from clear. In this work, we take a step towards understanding it by providing a model of how the loss landscape needs to behave topographically for LMC (or the lack thereof) to manifest. Concretely, we present a `mountainside and ridge' perspective that helps to neatly tie together different geometric features that can be spotted in the loss landscape along the training runs. We also complement this perspective by providing a theoretical analysis of the barrier height, for which we provide empirical support, and which additionally extends as a faithful predictor of layer-wise LMC. We close with a toy example that provides further intuition on how barriers arise in the first place, all in all, showcasing the larger aim of the work ---  to provide a working model of the landscape and its topography for the occurrence of LMC. 
\end{abstract}

\section{Introduction}
The loss landscape of over-parameterized neural networks, in general, and especially when taken as a whole, is undoubtedly non-convex. Yet, the confrontation with the still somewhat puzzling success of a local gradient based method to find generalizing solutions has led to a body of work that explores inherent regularity within optimization, such as through the lens of implicit bias~\citep{gunasekar2018characterizing,moroshko2020implicit}; or sources of structure and regularity within the landscape itself, like via its significant degeneracy~\citep{sagun2017empirical,singh2021analytic}, inherent symmetries~\citep{simsek2021geometry,entezari2021role}, or existence of monotonic paths during training~\citep{goodfellow2014qualitatively} as well as non-linear/linear paths between solutions~\citep{draxler2018essentially,garipov2018loss,frankle2020linear} --- all of which are likened to play a palliative role against the non-convexity. From this latter category, a particularly striking notion of regularity is that of linear mode connectivity~\cite{frankle2020linear}, where it has been observed that if two networks are made to share a common initial path of sufficient length (usually around $10\%$ of the whole training) and subsequently set apart on distinct paths from a `fork' (e.g., achieved via enforcing different orderings of the samples), they can nevertheless be connected with a linear path at convergence, along which the loss remains negligible. \looseness=-1

The particular interest of the research community in LMC can be attributed to, amongst other factors, the sheer simplicity with which a notion of convexity is demonstrated in the landscapes, as well as the practical implications it raises for model merging/fusion~\citep{singh2020model,ainsworth2022git}. As a result, there is a lot of research which tries to theoretically construct sets of paths~\citep{DBLP:journals/corr/abs-1912-10095,kuditipudi2020explaining,pmlr-v238-ferbach24a} to demonstrate LMC, as well as others which consider its extensions to convex hulls~\citep{yunis2022convexity}, feature~\citep{zhou2023going} and layer connectivity~\citep{adilova2023layer}, and increasingly, those that study its wider implications on generalization strategies~\citep{juneja2023linear}. However, amidst all these advances, our broader model of the loss landscape has seen little refinement. \textit{Hence, \textbf{our objective} is to precisely take a step towards this, by providing a model of the landscape that allows for LMC (or lack thereof) and explicates the various observations in this context. \looseness=-1
}

\paragraph{A metaphor for LMC.} Let us momentarily engage in a metaphor that explains our model.\looseness=-1
\begin{quote}
Imagine going down from near the top of a mountain towards the valley with a friend. With the weather being extremely foggy and windy, your visibility is negligible and you both chose to descend by locally following the downward slope. However, after a while into your hike downwards, you can't spot your friend and realize that you might have lost them on the way. Confident that you could not have diverged for long, you continue to march down with the hope that you will find them at the valley floor without strenuous effort. It takes some walking once you are finally down, but you soon run into them and are relieved. However, as you both gaze upwards, you realize that thankfully you did not separate around the top of the mountain itself, for a deviation then would tantamount to you both being on the different sides of the mountain, with a non-trivial barrier in between.\looseness=-1 %
\end{quote}

\section{The Topography of Landscapes vis-à-vis LMC}
\textbf{Key Hypothesis.} While the above is merely a metaphor, and not free from flaws, the resemblance to LMC is hard to ignore, and we take some inspiration from it to arrive at a hypothesis. 
Namely, that the occurrence of LMC can be explained by models moving down a mountain side with numerous ridges that are present at varying heights. If one forks on top of a plateau or a wide ridge, the child models may move to either side of it, effectively resulting in a barrier that completely prevents linear connectivity. In contrast, if one forks a little later on one side of the higher ridge, models largely remain on the same side with the mountain slope pulling them downwards, and if separated, are only divided by a lower ridge implying a small barrier.

\begin{figure}[!h]
    \centering
    \includegraphics[width=\textwidth]{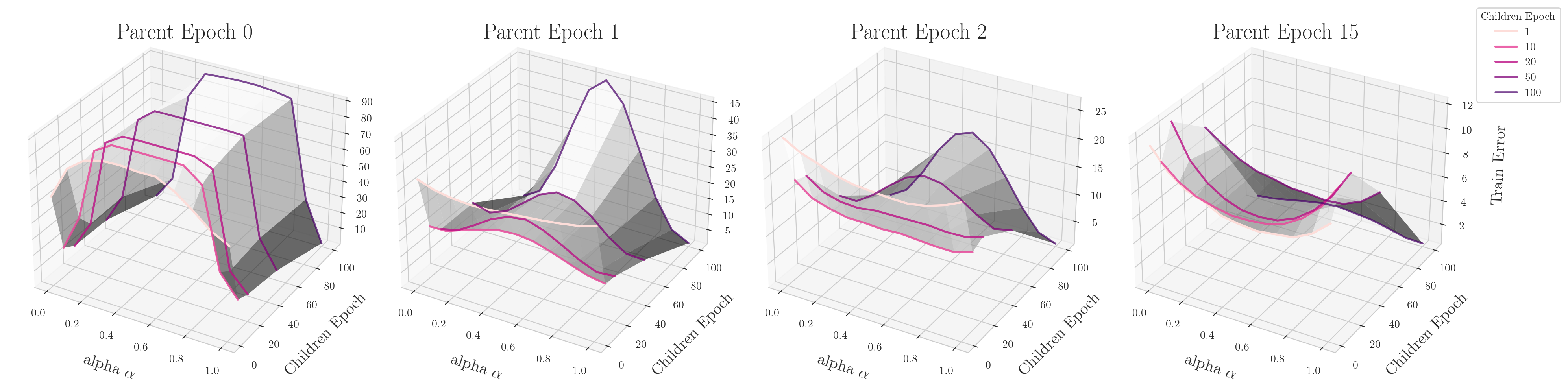}
    \caption{The evolution of training error curves, shown in different colors, when forked at different points. The barrier version (i.e., with the error of endpoints subtracted) is shown in Figure~\ref{fig:3d-barrier}.}
    \label{fig:resnet20-train-err}
\end{figure}

\paragraph{Empirical observations.} Let us directly take a look at the kind of cross-sectional loss landscape that appears between the child models (`siblings') when the parent model is forked at different points in training, and in particular, how this varies while training the child models. Figure~\ref{fig:resnet20-train-err} shows a plot of the training error on the line segment joining the siblings for the case of ResNet20 trained on CIFAR-10, mimicking the hyperparameter setup of~\citet{frankle2020linear} which is also detailed in Section~\ref{app:hyperparam}, when forked at epochs $0$ (i.e., initialization), $1, 2, 15$. The evolution of the training error is shown by laying them out in different planes, which also evokes how the siblings navigate the landscape.\looseness=-1

We observe that when the siblings are forked from the initialization, a wide ridge (i.e., considerable portion of the line segment is at high error) separates them throughout their training just as if they set out on different sides of the mountain. But when they are forked after a few (parent) epochs, they seem to be both going downward but are soon (in the span of $~10-20$ child epochs) separated by another ridge. This ridge although lower in height \textit{extends all the way until convergence}. Finally, if the children get forked much later, they seem to be simply descending downwards, and are separated by essentially a bump. These observations align neatly with our `mountain-and-ridge perspective' hypothesized above, and show most conspicuously, \textit{that barriers do not just show up right at convergence, but can be traced a long way before in the form of a ridge. }This is in contrast to the `static' final view of barriers portrayed in the literature, and as reproduced in Figure~\ref{fig:barrier-types}. Similar figures for different learning rate or without weight decay can be found in the Appendix~\ref{app:train-err-evols}.

\begin{figure}[ht!]
\vskip -0.1in
    \centering
    \subfigure[Sibling angles]{\label{fig:sib-angle-high}
\includegraphics[trim=5 10 5 20 , clip,width=0.3\textwidth]{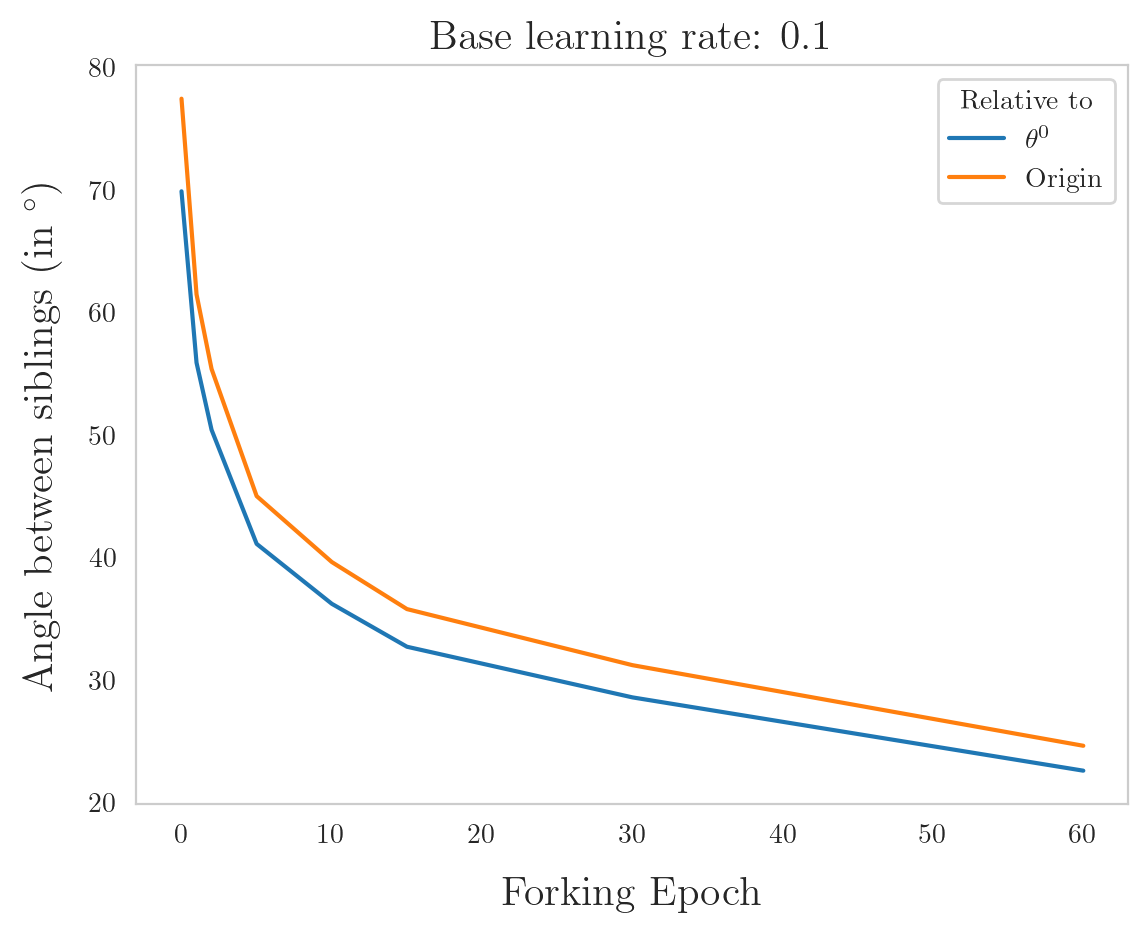}}
\hspace{15mm}
 \subfigure[Sibling solution planes]{
\includegraphics[trim=5 10 5 20 , clip,width=0.32\textwidth]{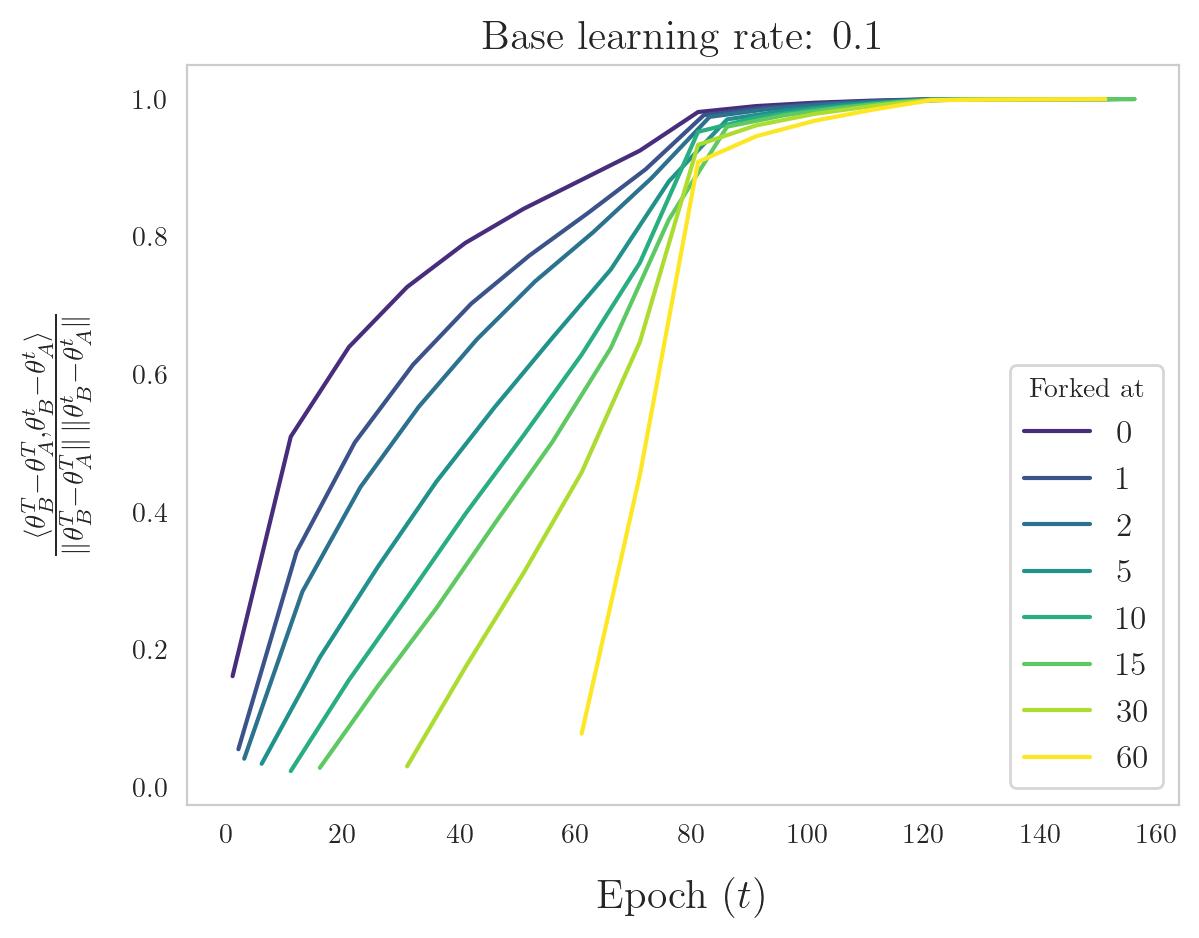}}
 \caption{The angle between sibling solutions (in degrees) as well as the determination of sibling solution planes for different forking epochs.}
 \label{fig:sibling-angles}
\end{figure}

\begin{wrapfigure}{R}{0.38\textwidth}
\vskip -0.2 in
\centering
\includegraphics[trim=5 10 5 10, clip, width=0.38\textwidth]{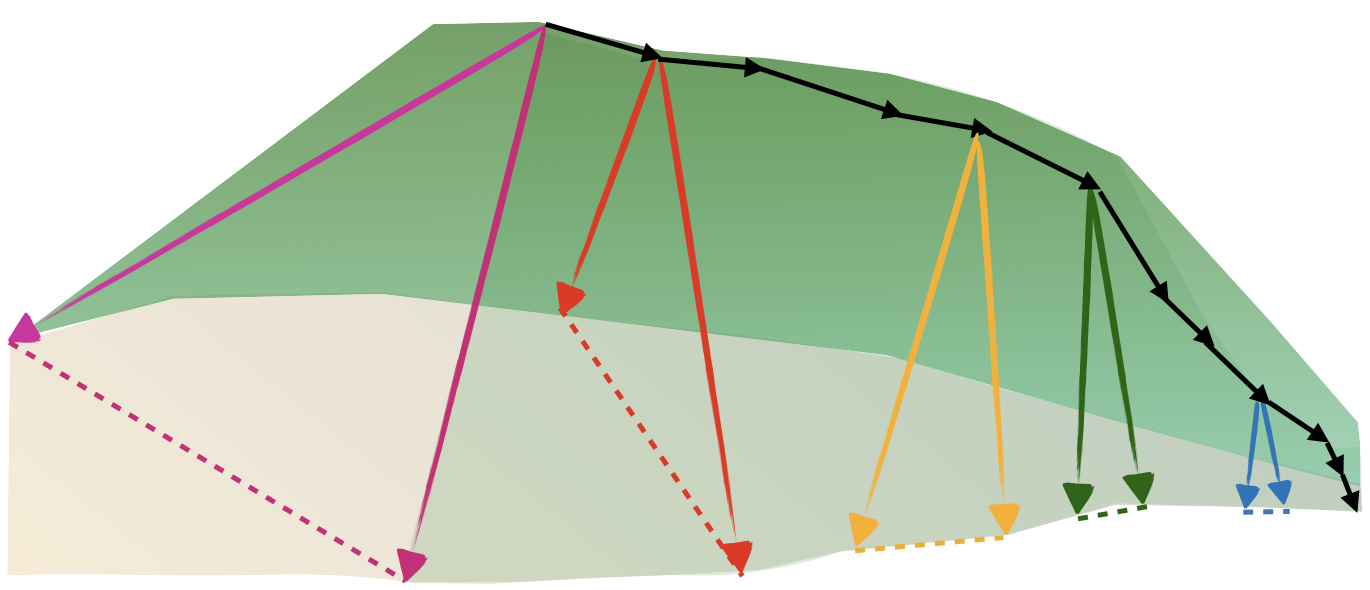}
 \caption{A sketch of different forks.}
  \label{fig:sketch}
\end{wrapfigure}

\textbf{Sibling angles and solution planes. }To further probe our hypothesis, we measure, for different forking points, the angle between the siblings $\soln{1},\soln{2}$ formed at a base point $\btheta_{\text{base}}$, which is either the origin or the respective forking point itself. More concretely, this amounts to measuring $\arccos(\langle\soln{1}-\btheta_{\text{base}}, \soln{2}-\btheta_{\text{base}}\rangle/\|\soln{1}-\btheta_{\text{base}}\|\|\soln{2}-\btheta_{\text{base}}\|)$ and is shown in Figure~\ref{fig:sibling-angles}(a). We observe that, regardless of the choice of the base point, the angle between earlier siblings is larger than that for later siblings, which is in line with our view, as shown pictorially in Figure~\ref{fig:sketch} that the earlier ridges lead to more lateral or cross-sectional separation than later ridges.

In a similar vein, we check, for different forking points, how early is the sibling solution plane determined, as measured by cosine similarity between sibling difference at step $t$ of their training, $\btheta^t_1 - \btheta^t_2$ and the final sibling difference $\soln{1}-\soln{2}$. The results are shown in Figure~\ref{fig:sibling-angles}(b), where we find that the siblings forked earlier have their solution plane determined most quickly than the later ones, which would occur if the former were indeed traversing on different sides of the planes and their cross-section largely determined rapidly into training.

Lastly, we also carry out a similar experiment when the models are trained with a lower learning rate, and the results for which we can be found in Figure~\ref{fig:sibling-low-lr}. The angles between siblings are even less and the sibling solution plane gets determined even faster, further corroborating our hypothesis.

\section{Barrier Analysis}
In the previous section, we saw that the various geometrical observations about the sibling solutions, across various forking points, fit neatly within the prescribed mountain-ridge hypothesis. Now, we would like to see if we can provide a model for the height of the ridge, and in particular, the eventual barrier that separates the sibling solutions.

Let us define a notion of barrier curve between the final solutions as, $\barrier{\alpha; \btheta_1, \btheta_2}= %
\Loss\big((1-\alpha)\btheta_1 + \alpha \btheta_2\big) - \left\lbrack(1-\alpha)\,\Loss(\btheta_1) + \alpha\,\Loss(\btheta_1)\right\rbrack$, parameterized for $\alpha\in\lbrack0,1\rbrack$. Usually, what is reported as the barrier is the maximum value~\citep{entezari2021role} of the barrier curve, and for the theoretical analysis we will consider the entire barrier curve. 
Further, the barrier curve is effectively a notion of non-convexity between the line segment joining two networks, which we refer to as `cross-sectional' non-convexity. If the (maximum) barrier comes out as negative, it implies convexity on this line-segment. Below, we provide an analysis that models the barrier up to the second-order in the distance between them.\looseness=-1
\begin{mdframed}[leftmargin=1mm,
    skipabove=1mm, 
    skipbelow=-1mm, 
    backgroundcolor=gray!10,
    linewidth=0pt,
    leftmargin=-1mm,
    rightmargin=-1mm,
    innerleftmargin=2mm,
    innerrightmargin=2mm,
    innertopmargin=1mm,
innerbottommargin=1mm]
\begin{proposition1}\label{theorem:lmc-barrier}
The loss barrier curve when linearly interpolating the final child networks $\soln{1}$ and $\soln{2}$ with weights $1-\alpha$ and $\alpha$ respectively, is given by\vspace{-3mm}
\begin{equation}\label{eq:barrier1}
    \barrier{\alpha; \soln{1}, \soln{2}}= \frac{\alpha(1-\alpha)}{2} (\soln{2} - \soln{1})^\top  \, \big(\alpha \nabla_\btheta^2 \Loss(\soln{1}) + (1-\alpha) \nabla_\btheta^2 \Loss(\soln{2}) \big)\, (\soln{2} - \soln{1})  +\mathcal{O}(\|\soln{2}-\soln{1}\|^3)
\end{equation}
\end{proposition1}
\end{mdframed}

The proof is located in the appendix~\ref{app:omitted-proofs}. From the form of the barrier, we make the following observations. (a) We see that the barrier increases if the two child networks solutions are more distant, but more precisely, the metric is with respect to the geometry of the convex-combination of their Hessians. (b) Since the networks are at convergence, the Hessian will be positive semi-definite by second-order optimality condition, the predicted barrier will always be non-negative though it might be small, unless higher-order terms kick in. (c) If we assume that the form of the local curvatures is similar (this doesn't mean they have to be the same, but more that the dominant eigenvectors are aligned), then we can approximate the barrier by $
    \barrier{\alpha; \soln{1}, \soln{2}}\approx \frac{\alpha(1-\alpha)}{2} (\soln{2} - \soln{1})^\top  \, \cdot \, \nabla_\btheta^2 \Loss(\soln{1}) \, \cdot \, (\soln{2} - \soln{1})$
, and from where it is easy to see that it is maximized for $\alpha=1/2$, as often seen in practice.

\begin{figure}[ht!]
    \centering
    \subfigure[\textcolor{blue}{Predicted} vs \textcolor{orange}{Actual} Barriers]{
	\includegraphics[width=0.32\textwidth]{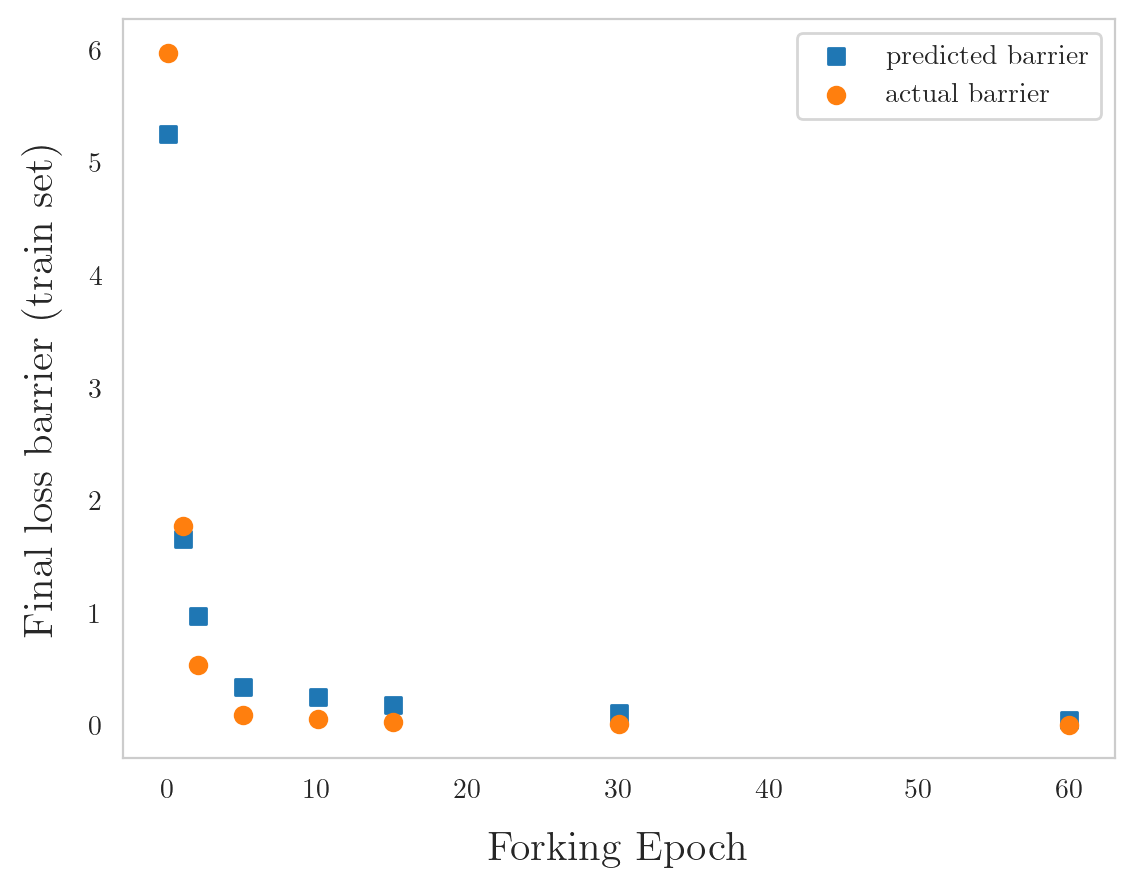}}
 \subfigure[Sibling Distances]{
    	\includegraphics[width=0.37\textwidth]{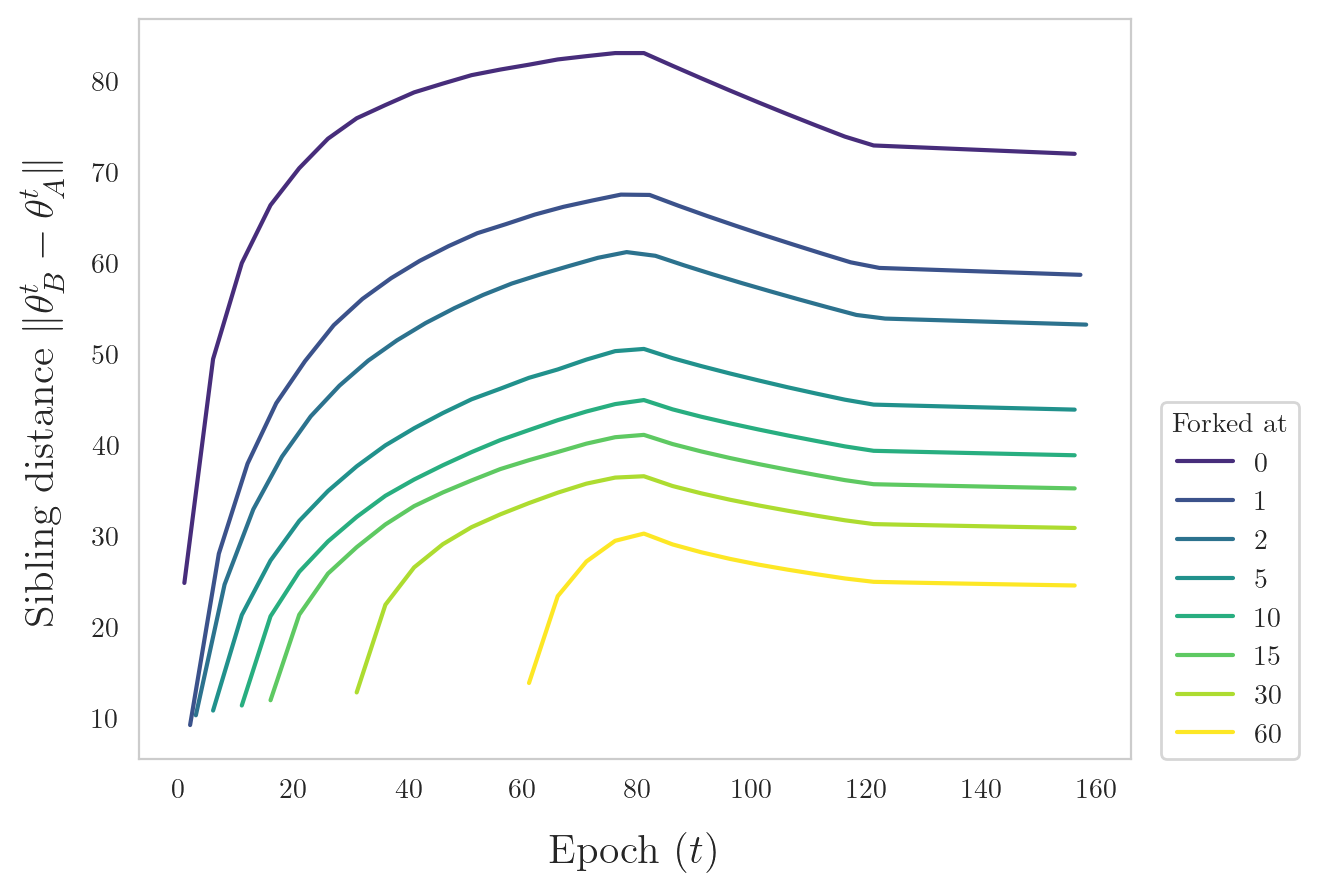}
 }
    \caption{\textit{Left:} Barrier predictions versus actual barriers for final solutions obtained by forking the parent trajectory at different epochs in its training. \textit{Right: }The evolution of the distance between sibling models for different forking points.%
    }
    \label{fig:forks-dists1}
\end{figure}

Next, in Figure~\ref{fig:forks-dists1}(a) we compare the fidelity of our barrier predictions with what is observed empirically. We find that these predictions fall in decent ballpark of the actual barriers, even though the distance between the models under consideration is significant as evident from~\ref{fig:forks-dists1}(b), and where the higher-order terms should come into the picture. %

 \begin{wrapfigure}{R}{0.46\textwidth}
\vskip -0.4in
    \centering
    \subfigure[\small{loss barrier}]{
        \label{fig:loss_1}
		\includegraphics[trim=5 7 6 20, clip,width=0.2\textwidth]{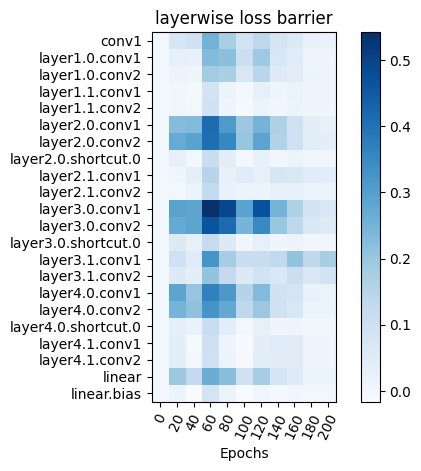}}\hspace{2mm}
  \subfigure[\small{predicted barrier}]{\label{fig:loss_2}
    \includegraphics[trim=5 7 6 20, clip,width=0.2\textwidth]{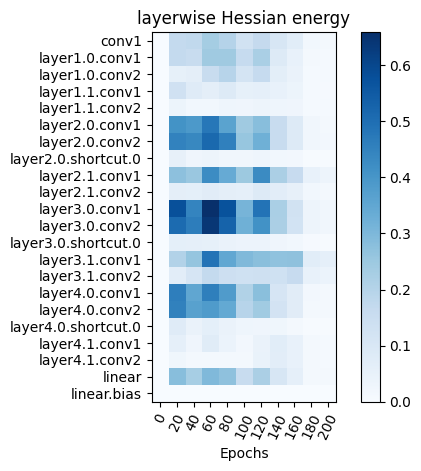}}
    \caption{\textbf{(Left)} The layerwise loss barrier as per~\cite{adilova2023layer}. \textbf{(Right):} Layerwise predicted barrier, as per Proposition~\ref{theorem:llmc-barrier}, during the course of training ResNet18 (see Appendix~\ref{app:layerwise-details} for details).}
    \label{fig:layerwise-weighted}
\end{wrapfigure}

\vspace{1mm}
\textbf{Extension to layerwise LMC.} \citet{adilova2023layer} suggested a notion of layerwise LMC, where only the parameters of single individual layers are linearly interpolated. Layerwise LMC can thus provide a more fine-grained view of (non)convexity and has also been observed to hold even when the entire networks may not be linearly connected. We extend our barrier analysis to the layerwise case in Proposition~\ref{theorem:llmc-barrier}, and Figure~\ref{fig:layerwise-weighted} shows that these predictions provide a rather faithful match with the true layerwise barriers. Besides, Appendix~\ref{app:layerwise-more} contains additional results, where we find that our barrier predictions capture the magnitude of the actual barriers, and that both the Hessian term in the barrier expression above as well as the distance between the parameters are important to model the barrier closely.\looseness=-1

\begin{wrapfigure}{R}{0.32\textwidth}
\vskip -0.5 in
    \centering    
    \subfigure[two solutions]{
    \includegraphics[width=0.3\textwidth]{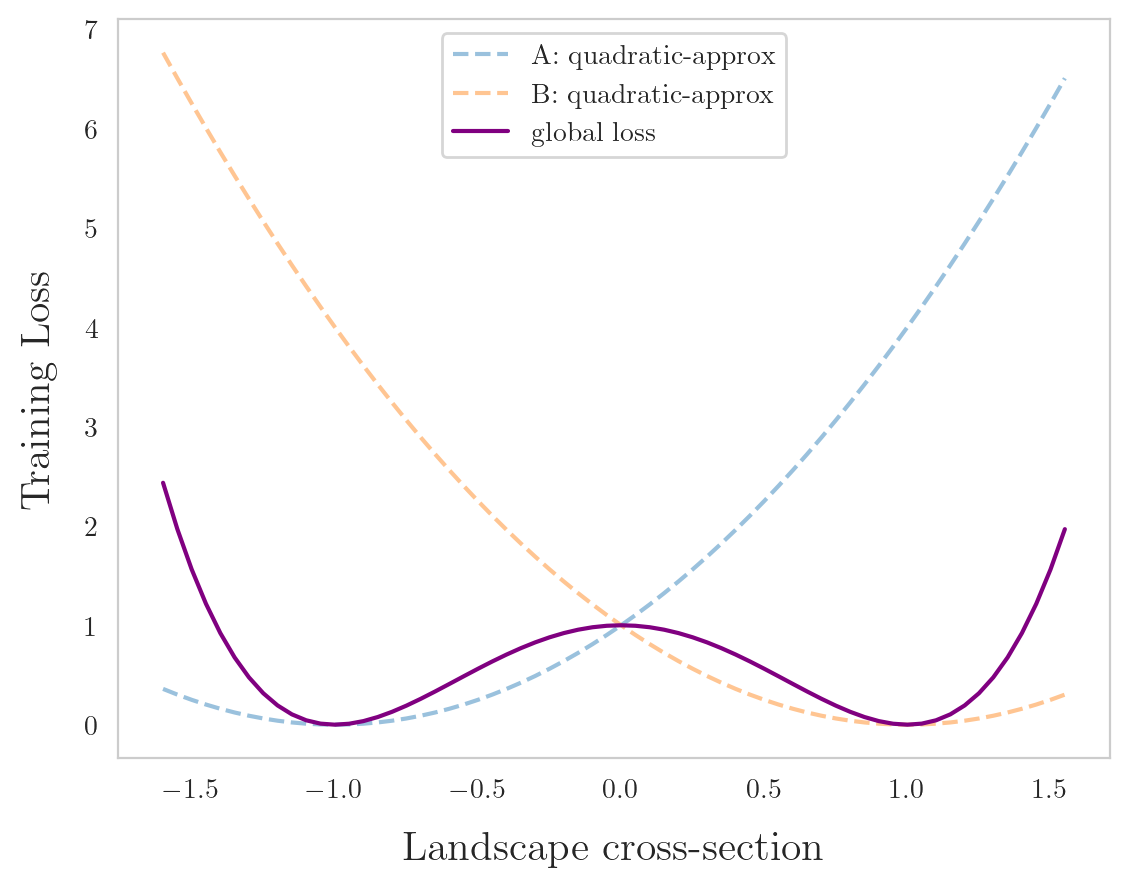}}
    \subfigure[multiple solutions]{\includegraphics[width=0.3\textwidth]{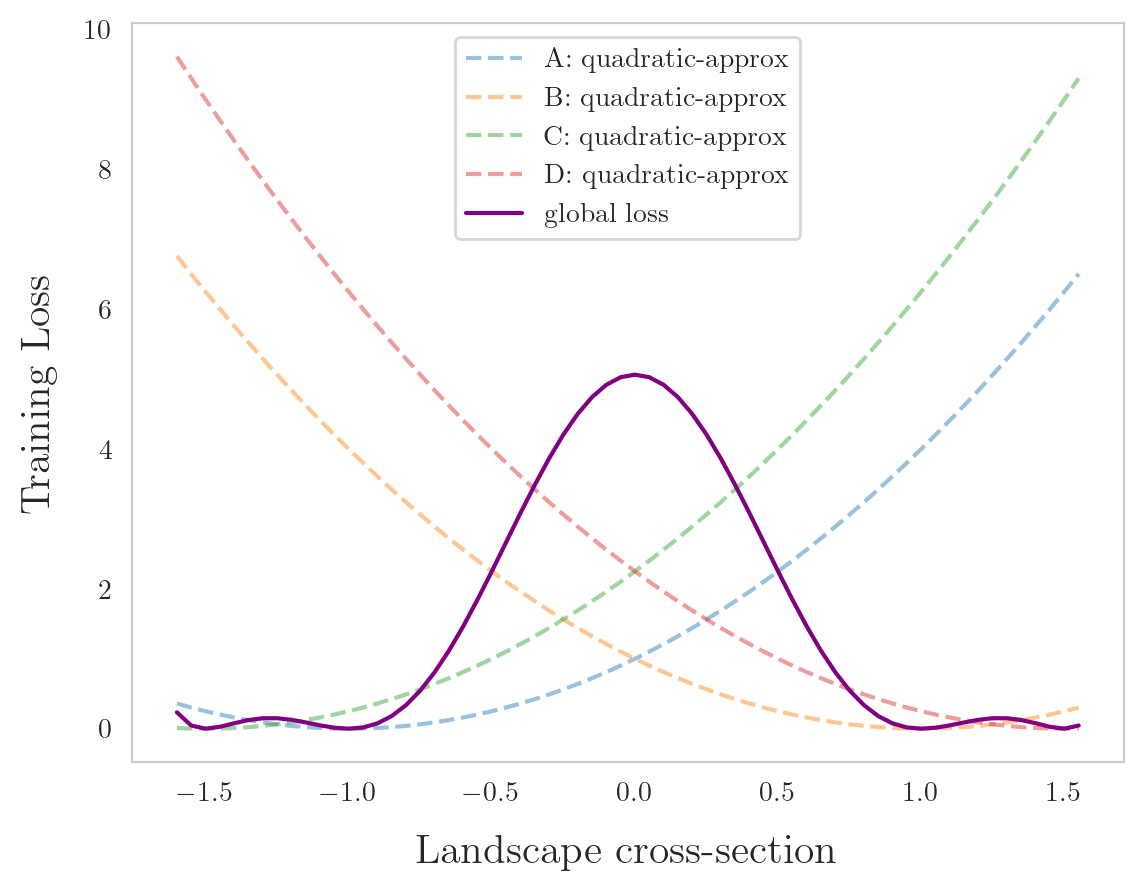}}
    \caption{Toy examples: loss landscape cross-section.}
    \label{fig:toy-example}
\end{wrapfigure}
\vspace{-2mm}
\section{Discussion}
\textbf{Summary. }To sum up, we provide a new unified perspective to think about LMC, inspired by a mountainside-ridge view, that explains various aspects of LMC such as how early and later forked solutions are situated in the landscape, and how they can be disconnected or connected. We also present a theoretical analysis which indicates that the extent of connectivity depends on the sibling distance and the local sibling curvature, and demonstrate that it can empirically provide decent barrier estimates.

\vspace{1mm}
\textbf{A retrospective on barriers.} In an alternate way, let us further consolidate our intuition on how barriers can arise at all. Consider a 1-dimensional toy example below, with two minima, one at $\theta^\ast_1=-1$ and the other at $\theta^\ast_2=1$. Locally, %
since both are
valid local minima, a quadratic approximation explains the loss surface, i.e., \textcolor{blue}{$(\theta-\theta^\ast_1)^2$} and \textcolor{orange}{$(\theta-\theta^\ast_2)^2$} respectively.
Then the simplest model of the landscape that would jointly have both these local minima is simply the product of the two quadratic approximations, i.e., \textcolor{violet}{$(\theta-\theta^\ast_1)^2\,\cdot\,(\theta-\theta^\ast_2)^2$}. Notice how this joint loss model has a barrier in between the two minima, as shown in Figure~\ref{fig:toy-example}(a). In fact, since we don't just have two solutions, we can extend the above example to include two further minima at $-1.5, 1.5$ as shown in~Figure~\ref{fig:toy-example}(b), and we can imagine the larger idea of how a hierarchy of barriers might emerge, with the barrier between distant solutions being higher --- something which the Proposition~\ref{theorem:lmc-barrier} also shows, but which additionally accounts for the local curvatures. %

\vspace{1mm}
\textbf{Future Work.} There are still several interesting questions that require further study: (a) Is there an early geometric indicator which can predict the extent of the final barrier? (b) Can we go beyond the second-order model of the barrier, while being efficient, and further refine the barrier predictions? (c) Given the rapidly determined cross-sectional direction, can it be utilized for model fusion without having to train all the child networks until convergence?

\bibliography{sample}

\begin{thebibliography}{19}
\providecommand{\natexlab}[1]{#1}
\providecommand{\url}[1]{\texttt{#1}}
\expandafter\ifx\csname urlstyle\endcsname\relax
  \providecommand{\doi}[1]{doi: #1}\else
  \providecommand{\doi}{doi: \begingroup \urlstyle{rm}\Url}\fi

\bibitem[Adilova et~al.(2023)Adilova, Andriushchenko, Kamp, Fischer, and
  Jaggi]{adilova2023layer}
Linara Adilova, Maksym Andriushchenko, Michael Kamp, Asja Fischer, and Martin
  Jaggi.
\newblock Layer-wise linear mode connectivity.
\newblock In \emph{The Twelfth International Conference on Learning
  Representations}, 2023.

\bibitem[Ainsworth et~al.(2022)Ainsworth, Hayase, and
  Srinivasa]{ainsworth2022git}
Samuel Ainsworth, Jonathan Hayase, and Siddhartha Srinivasa.
\newblock Git re-basin: Merging models modulo permutation symmetries.
\newblock In \emph{The Eleventh International Conference on Learning
  Representations}, 2022.

\bibitem[Draxler et~al.(2018)Draxler, Veschgini, Salmhofer, and
  Hamprecht]{draxler2018essentially}
Felix Draxler, Kambis Veschgini, Manfred Salmhofer, and Fred Hamprecht.
\newblock Essentially no barriers in neural network energy landscape.
\newblock In \emph{International conference on machine learning}, pages
  1309--1318. PMLR, 2018.

\bibitem[Entezari et~al.(2021)Entezari, Sedghi, Saukh, and
  Neyshabur]{entezari2021role}
Rahim Entezari, Hanie Sedghi, Olga Saukh, and Behnam Neyshabur.
\newblock The role of permutation invariance in linear mode connectivity of
  neural networks.
\newblock In \emph{International Conference on Learning Representations}, 2021.

\bibitem[Ferbach et~al.(2024)Ferbach, Goujaud, Gidel, and
  Dieuleveut]{pmlr-v238-ferbach24a}
Damien Ferbach, Baptiste Goujaud, Gauthier Gidel, and Aymeric Dieuleveut.
\newblock Proving linear mode connectivity of neural networks via optimal
  transport.
\newblock In Sanjoy Dasgupta, Stephan Mandt, and Yingzhen Li, editors,
  \emph{Proceedings of The 27th International Conference on Artificial
  Intelligence and Statistics}, volume 238 of \emph{Proceedings of Machine
  Learning Research}, pages 3853--3861. PMLR, 02--04 May 2024.
\newblock URL \url{https://proceedings.mlr.press/v238/ferbach24a.html}.

\bibitem[Frankle et~al.(2020)Frankle, Dziugaite, Roy, and
  Carbin]{frankle2020linear}
Jonathan Frankle, Gintare~Karolina Dziugaite, Daniel Roy, and Michael Carbin.
\newblock Linear mode connectivity and the lottery ticket hypothesis.
\newblock In \emph{International Conference on Machine Learning}, pages
  3259--3269. PMLR, 2020.

\bibitem[Garipov et~al.(2018)Garipov, Izmailov, Podoprikhin, Vetrov, and
  Wilson]{garipov2018loss}
Timur Garipov, Pavel Izmailov, Dmitrii Podoprikhin, Dmitry~P Vetrov, and
  Andrew~G Wilson.
\newblock Loss surfaces, mode connectivity, and fast ensembling of dnns.
\newblock \emph{Advances in neural information processing systems}, 31, 2018.

\bibitem[Goodfellow et~al.(2015)Goodfellow, Vinyals, and
  Saxe]{goodfellow2014qualitatively}
Ian~J Goodfellow, Oriol Vinyals, and Andrew~M Saxe.
\newblock Qualitatively characterizing neural network optimization problems.
\newblock \emph{ICLR}, 2015.

\bibitem[Gunasekar et~al.(2018)Gunasekar, Lee, Soudry, and
  Srebro]{gunasekar2018characterizing}
Suriya Gunasekar, Jason Lee, Daniel Soudry, and Nathan Srebro.
\newblock Characterizing implicit bias in terms of optimization geometry.
\newblock In \emph{International Conference on Machine Learning}, pages
  1832--1841. PMLR, 2018.

\bibitem[Juneja et~al.(2023)Juneja, Bansal, Cho, Sedoc, and
  Saphra]{juneja2023linear}
Jeevesh Juneja, Rachit Bansal, Kyunghyun Cho, João Sedoc, and Naomi Saphra.
\newblock Linear connectivity reveals generalization strategies, 2023.

\bibitem[Kuditipudi et~al.(2020)Kuditipudi, Wang, Lee, Zhang, Li, Hu, Arora,
  and Ge]{kuditipudi2020explaining}
Rohith Kuditipudi, Xiang Wang, Holden Lee, Yi~Zhang, Zhiyuan Li, Wei Hu,
  Sanjeev Arora, and Rong Ge.
\newblock Explaining landscape connectivity of low-cost solutions for
  multilayer nets, 2020.

\bibitem[Moroshko et~al.(2020)Moroshko, Gunasekar, Woodworth, Lee, Srebro, and
  Soudry]{moroshko2020implicit}
Edward Moroshko, Suriya Gunasekar, Blake Woodworth, Jason~D. Lee, Nathan
  Srebro, and Daniel Soudry.
\newblock Implicit bias in deep linear classification: Initialization scale vs
  training accuracy, 2020.

\bibitem[Sagun et~al.(2017)Sagun, Evci, Guney, Dauphin, and
  Bottou]{sagun2017empirical}
Levent Sagun, Utku Evci, V~Ugur Guney, Yann Dauphin, and Leon Bottou.
\newblock Empirical analysis of the hessian of over-parametrized neural
  networks.
\newblock \emph{arXiv preprint arXiv:1706.04454}, 2017.

\bibitem[Shevchenko and Mondelli(2019)]{DBLP:journals/corr/abs-1912-10095}
Alexander Shevchenko and Marco Mondelli.
\newblock Landscape connectivity and dropout stability of {SGD} solutions for
  over-parameterized neural networks.
\newblock \emph{CoRR}, abs/1912.10095, 2019.
\newblock URL \url{http://arxiv.org/abs/1912.10095}.

\bibitem[Simsek et~al.(2021)Simsek, Ged, Jacot, Spadaro, Hongler, Gerstner, and
  Brea]{simsek2021geometry}
Berfin Simsek, Fran{\c{c}}ois Ged, Arthur Jacot, Francesco Spadaro, Cl{\'e}ment
  Hongler, Wulfram Gerstner, and Johanni Brea.
\newblock Geometry of the loss landscape in overparameterized neural networks:
  Symmetries and invariances.
\newblock In \emph{International Conference on Machine Learning}, pages
  9722--9732. PMLR, 2021.

\bibitem[Singh and Jaggi(2020)]{singh2020model}
Sidak~Pal Singh and Martin Jaggi.
\newblock Model fusion via optimal transport.
\newblock \emph{Advances in Neural Information Processing Systems},
  33:\penalty0 22045--22055, 2020.

\bibitem[Singh et~al.(2021)Singh, Bachmann, and Hofmann]{singh2021analytic}
Sidak~Pal Singh, Gregor Bachmann, and Thomas Hofmann.
\newblock Analytic insights into structure and rank of neural network hessian
  maps.
\newblock \emph{Advances in Neural Information Processing Systems},
  34:\penalty0 23914--23927, 2021.

\bibitem[Yunis et~al.(2022)Yunis, Patel, Savarese, Vardi, Frankle, Walter,
  Livescu, and Maire]{yunis2022convexity}
David Yunis, Kumar~Kshitij Patel, Pedro Henrique~Pamplona Savarese, Gal Vardi,
  Jonathan Frankle, Matthew Walter, Karen Livescu, and Michael Maire.
\newblock On convexity and linear mode connectivity in neural networks.
\newblock In \emph{OPT 2022: Optimization for Machine Learning (NeurIPS 2022
  Workshop)}, 2022.

\bibitem[Zhou et~al.(2023)Zhou, Yang, Yang, Yan, and Hu]{zhou2023going}
Zhanpeng Zhou, Yongyi Yang, Xiaojiang Yang, Junchi Yan, and Wei Hu.
\newblock Going beyond linear mode connectivity: The layerwise linear feature
  connectivity, 2023.

\end{thebibliography}
\clearpage
\appendix
\section{Omitted Proofs}\label{app:omitted-proofs}

\subsection{LMC barrier proof}
\begin{mdframed}[leftmargin=1mm,
    skipabove=1mm, 
    skipbelow=-1mm, 
    backgroundcolor=gray!10,
    linewidth=0pt,
    leftmargin=-1mm,
    rightmargin=-1mm,
    innerleftmargin=2mm,
    innerrightmargin=2mm,
    innertopmargin=0mm,
innerbottommargin=1mm]
\begin{repproposition1}{theorem:lmc-barrier}
The loss barrier when linearly interpolating the final child networks $\soln{1}$ and $\soln{2}$, forked from a common point $\btheta_0$, with weights $1-\alpha$ and $\alpha$ respectively, is given by
\begin{align*}\label{eq:barrier1-app}
    \barrier{\alpha; \soln{1}, \soln{2}}= \frac{\alpha(1-\alpha)}{2} (\soln{2} - \soln{1})^\top  \, \big(\alpha \nabla_\btheta^2 \Loss(\soln{1}) + (1-\alpha) \nabla_\btheta^2 \Loss(\soln{2}) \big)\, (\soln{2} - \soln{1})  +\mathcal{O}(\|\soln{2}-\soln{1}\|^3)
\end{align*}
\end{repproposition1}
\end{mdframed}
\begin{proof}

Using the Taylor series we have that, 
\begin{align}
    &\Loss\big((1-\alpha)\soln{1} + \alpha \soln{2}\big) = \Loss\big(\soln{1} + \alpha (\soln{2}-\soln{1})\big)\\
    &= \Loss(\soln{1}) + \alpha \nabla_\btheta \Loss(\soln{1})^\top (\soln{2} - \soln{1}) + \frac{\alpha^2}{2} (\soln{2} - \soln{1})^\top \, \nabla_\btheta^2 \Loss(\soln{1}) \, (\soln{2} - \soln{1}) + \mathcal{O}(\|\soln{2} - \soln{1}\|^3)\\
    &= \Loss(\soln{1}) + \frac{\alpha^2}{2} (\soln{2} - \soln{1})^\top \, \nabla_\btheta^2 \Loss(\soln{1}) \, (\soln{2} - \soln{1})  + \mathcal{O}(\|\soln{2} - \soln{1}\|^3)\label{eq:loss1}
\end{align}
The last line follows from the fact that at  the optimum, $\nabla_\btheta\Loss(\soln{1})=\mathbf{0}$ and $\nabla_\btheta\Loss(\soln{2})=\mathbf{0}$.
Likewise, we can repeat the above steps with $\soln{2}$ as the center of the Taylor series expansion, which results in:
\begin{align}
    \Loss\big((1-\alpha)\soln{1} + \alpha \soln{2}\big) &= \Loss\big(\soln{2} - (1-\alpha)(\soln{2} - \soln{1})\big)\\ \label{eq:loss2}
    &= \Loss(\soln{2}) + \frac{(1-\alpha)^2}{2} (\soln{2} - \soln{1})^\top  \, \nabla_\btheta^2 \Loss(\soln{2}) \, (\soln{2} - \soln{1}) + \mathcal{O}(\|\soln{2} - \soln{1}\|^3)
\end{align}
Multiplying\footnote{Note while the factors with which the equations are multiplied are mathematically convenient to obtain the definition of the barrier, they also make sense in that when $\alpha$ is small, ~\eqref{eq:loss1} will be 
 a more accurate model of the loss at the interpolation and gets a higher weight of $1-\alpha$, and likewise when $\alpha$ is large (or $1-\alpha$ is small)~\eqref{eq:loss2} is weighed in more to yield an accurate model of the loss at the interpolated point.}~\eqref{eq:loss1} by $1-\alpha$ and~\eqref{eq:loss2} by $\alpha$, and then adding them yields:
\begin{align}
    \Loss\big((1-\alpha)\soln{1} + \alpha \soln{2}\big) 
    &= (1-\alpha) \Loss(\soln{1})  + \alpha \Loss(\soln{2})  \\
    &+\frac{\alpha(1-\alpha)}{2} (\soln{2} - \soln{1})^\top  \, \big(\alpha \nabla_\btheta^2 \Loss(\soln{1}) + (1-\alpha) \nabla_\btheta^2 \Loss(\soln{2}) \big)\, (\soln{2} - \soln{1}) \\
    &+ \mathcal{O}(\|\soln{2} - \soln{1}\|^3)
\end{align}
Rearranging the terms we get that the following expression for the barrier between the two solutions:
\begin{align}\label{eq:barrier1-proof}
    \barrier{\alpha}= \frac{\alpha(1-\alpha)}{2} (\soln{2} - \soln{1})^\top  \, \big(\alpha \nabla_\btheta^2 \Loss(\soln{1}) + (1-\alpha) \nabla_\btheta^2 \Loss(\soln{2}) \big)\, (\soln{2} - \soln{1}) + \mathcal{O}(\|\soln{2} - \soln{1}\|^3)
\end{align}
\QEDA\end{proof}
\clearpage
\subsection{Layerwise LMC barrier proof}

\begin{mdframed}[leftmargin=1mm,
    skipabove=1mm, 
    skipbelow=-1mm, 
    backgroundcolor=gray!10,
    linewidth=0pt,
    leftmargin=-1mm,
    rightmargin=-1mm,
    innerleftmargin=2mm,
    innerrightmargin=2mm,
    innertopmargin=0mm,
innerbottommargin=1mm]
\begin{proposition1}\label{theorem:llmc-barrier}
The loss barrier when linearly interpolating only the layer $\ell$ parameters of the final child networks $\soln{1}$ and $\soln{2}$, forked from a common point $\btheta_0$, with weights $1-\alpha$ and $\alpha$ respectively, is
\begin{align*}\label{eq:barrier1-layerwise}
   \mathcal{B}_{\ell}(\alpha) = \frac{\alpha(1-\alpha)}{2} \Delta{\btheta^\ast\lbrack\ell\rbrack}^\top \, \big(\alpha\,\nabla_\btheta^2 \Loss(\soln{1}\lbrack\ell\rbrack)  + (1-\alpha)\,\nabla_\btheta^2 \Loss(\soln{2}\lbrack\ell\rbrack)  \big)\, \Delta\btheta^\ast\lbrack\ell\rbrack+ \mathcal{O}(\| \Delta\btheta^\ast\lbrack \ell \rbrack\|^3)
\end{align*}
\end{proposition1}
\end{mdframed}
\begin{proof}

Instead of considering the line-segment between all the network parameters, the work of ~\citet{adilova2023layer} considers the case where only a single layer's parameters are interpolated into another model. Let us now repeat the analysis, assuming we interpolate only with the layer $\ell$ of the second network into the first. If we designate the layer-wise parameters by the superscript as $\btheta\lbrack\ell\rbrack$, this amounts to:
$$
{\btheta}_{2\rightarrow 1}\lbrack\ell\rbrack:= (1-\alpha)\,\cdot\,\soln{1}\lbrack\ell\rbrack + \alpha \, \cdot \soln{2}\lbrack\ell\rbrack\,,\quad \text{and}\quad {\btheta}_{2\rightarrow 1}\lbrack\ell'\rbrack:= \soln{1}\lbrack\ell'\rbrack \, \quad \forall \, \ell'\neq \ell
$$
Likewise, we can consider the interpolation of the layer $\ell$ of the first network into the second, yielding the parameters $\btheta_{1\rightarrow 2}$ defined as follows:
$$
{\btheta}_{1\rightarrow 2}\lbrack\ell\rbrack:= (1-\alpha)\,\cdot\,\soln{1}\lbrack\ell\rbrack + \alpha \, \cdot \soln{2}\lbrack\ell\rbrack\,,\quad \text{and}\quad {\btheta}_{1\rightarrow 2}\lbrack\ell'\rbrack:= \soln{2}\lbrack\ell'\rbrack \, \quad \forall \, \ell'\neq \ell
$$
In other words, $\btheta_{2\rightarrow 1}$ and $\btheta_{1\rightarrow 2}$ only differ in terms of where to take the parameters for other layers, whether from $\soln{1}$ or from $\soln{2}$ respectively.

Furthermore, let $\Pm_{\ell}\in\mathbb{R}^{p\times p}$ stand for the diagonal matrix, with $i^{\text{th}}$ entry is $\dsone\lbrace \btheta_i \in \btheta\lbrack\ell\rbrack \rbrace$, i.e.,  which contains $1$ at the index which corresponds to the parameter from layer $\ell$, represented by $\btheta\lbrack\ell\rbrack$, and $0$ elsewhere. Then we can write the above parameter interpolations more succinctly as:
$$
    \btheta_{2\rightarrow 1}^{(\ell)} = \soln{1} \, + \, \alpha \, \cdot \, \Pm_{\ell} \cdot \Delta \btheta^\ast \,,\quad \text{and}\quad 
    \btheta_{1\rightarrow 2}^{(\ell)} = \soln{2} \,-\, (1-\alpha) \, \cdot \, \Pm_{\ell} \cdot \Delta\btheta^\ast 
$$
where, $\Delta\btheta^\ast := \soln{2}-\soln{1}$. Next, we apply a second-order Taylor series to approximate the loss at these interpolations.
\begin{align}
    \Loss(\btheta_{2\rightarrow1}^{(\ell)}) 
    &= \Loss(\soln{1}) + \alpha \nabla_\btheta \Loss(\soln{1})^\top \Pm_\ell \Delta\btheta^\ast + \frac{\alpha^2}{2}  \Delta{\btheta^\ast}^\top \, \Pm_\ell\, \nabla_\btheta^2 \Loss(\soln{1})  \, \Pm_\ell \, \Delta\btheta^\ast+ \mathcal{O}(\|\Pm_\ell\, \Delta\btheta^\ast\|^3)\\
    &=\Loss(\soln{1}) + \frac{\alpha^2}{2}  \Delta{\btheta^\ast}^\top \, \Pm_\ell\, \nabla_\btheta^2 \Loss(\soln{1})  \, \Pm_\ell \, \Delta\btheta^\ast+ \mathcal{O}(\|\Pm_\ell\, \Delta\btheta^\ast\|^3)\\
    & = \Loss(\soln{1}) + \frac{\alpha^2}{2}  \Delta{\btheta^\ast\lbrack\ell\rbrack}^\top \, \nabla_\btheta^2 \Loss(\soln{1}\lbrack\ell\rbrack)  \, \Delta\btheta^\ast\lbrack\ell\rbrack + \mathcal{O}(\| \Delta\btheta^\ast\lbrack \ell \rbrack\|^3)\,,\label{eq:layerwise21}
\end{align}
where the second term indicates the quadratic form with the Hessian at layer $\ell$ multiplied by the difference in parameters on that layer $\ell$ (i.e., $\Delta\btheta^\ast\lbrack\ell\rbrack$) and the first-order terms go away since $\soln{1}$ is a stationary point. Similarly, we get for the other interpolation:
\begin{align}\label{eq:layerwise12}
    \Loss(\btheta_{1\rightarrow 2}^{(\ell)}) &= \Loss(\soln{2}) + \frac{(1-\alpha)^2}{2}  \Delta{\btheta^\ast\lbrack\ell\rbrack}^\top \, \nabla_\btheta^2 \Loss(\soln{2}\lbrack\ell\rbrack)  \, \Delta\btheta^\ast\lbrack\ell\rbrack+ \mathcal{O}(\| \Delta\btheta^\ast\lbrack \ell \rbrack\|^3)\,,
\end{align}
Multiplying~\eqref{eq:layerwise21} by $1-\alpha$ and~\eqref{eq:layerwise12} by $\alpha$, we get:
\begin{align}
    (1-\alpha) \Loss(\btheta_{2\rightarrow1}) + \alpha \Loss(\btheta_{1\rightarrow2}) &= (1-\alpha) \Loss(\soln{1}) + \alpha \Loss(\soln{2}) \\
    &+ \frac{\alpha(1-\alpha)}{2} \Delta{\btheta^\ast\lbrack\ell\rbrack}^\top \, \big(\alpha\,\nabla_\btheta^2 \Loss(\soln{1}\lbrack\ell\rbrack)  + (1-\alpha)\,\nabla_\btheta^2 \Loss(\soln{2}\lbrack\ell\rbrack)  \big)\, \Delta\btheta^\ast\lbrack\ell\rbrack\\
    &  + \mathcal{O}(\| \Delta\btheta^\ast\lbrack \ell \rbrack\|^3)\,,
\end{align}
If we now define the layer-wise barrier as, $$\mathcal{B}_{\ell}(\alpha):= (1-\alpha) \Loss(\btheta_{2\rightarrow1}^{(\ell)}) + \alpha \Loss(\btheta_{1\rightarrow2}^{(\ell)}) - \lbrack(1-\alpha) \Loss(\soln{1}) + \alpha \Loss(\soln{2})\rbrack\,,$$
the above analysis reduces to the following expression:
\begin{align}
    \mathcal{B}_{\ell}(\alpha) = \frac{\alpha(1-\alpha)}{2} \Delta{\btheta^\ast\lbrack\ell\rbrack}^\top \, \big(\alpha\,\nabla_\btheta^2 \Loss(\soln{1}\lbrack\ell\rbrack)  + (1-\alpha)\,\nabla_\btheta^2 \Loss(\soln{2}\lbrack\ell\rbrack)  \big)\, \Delta\btheta^\ast\lbrack\ell\rbrack+ \mathcal{O}(\| \Delta\btheta^\ast\lbrack \ell \rbrack\|^3)
\end{align}
\QEDA
\end{proof}

\paragraph{Remarks.} Now, making the approximation of similar layer-wise Hessians $\nabla_\btheta^2 \Loss(\soln{1}\lbrack\ell\rbrack)  \approx \nabla_\btheta^2 \Loss(\soln{2}\lbrack\ell\rbrack)$, we can further simplify the above expression to:
\begin{align}
    {\mathcal{B}_{\ell}}(\alpha) \approx \frac{\alpha(1-\alpha)}{2} \Delta{\btheta^\ast\lbrack\ell\rbrack}^\top \, \,\nabla_\btheta^2 \Loss(\soln{1}\lbrack\ell\rbrack)\,\, \Delta\btheta^\ast\lbrack\ell\rbrack\,,
\end{align}
which also attains its maximum value for $\alpha=\frac{1}{2}$. The above expression is similar to what we had before, except our parameter update here concerns only the layer $\ell$ and the Hessian is also thus for this layer only. Although, here we only require much weaker assumptions, as our parameter update is local to layer $\ell$ and similarity of the Hessian is being considered only with respect to the diagonal block corresponding to layer $\ell$.

 \paragraph{Generalization to arbitrary set of layers.} Notice that can be rewritten, at $\alpha=1/2$,  as
\begin{align}\label{eq:layerwisebarrier-gen}
    {\mathcal{B}}^\ast \approx \frac{1}{8} \sum_{\ell,\ell' \,\in\,\lbrack1,L\rbrack} \, \Delta{\btheta^\ast\lbrack\ell\rbrack}^\top \, \cdot \,\nabla_\btheta^2 \Loss(\soln{1}\lbrack\ell,\ell'\rbrack)\,\cdot \, \Delta\btheta^\ast\lbrack\ell'\rbrack=\sum_{\ell,\ell' \,\in\,\lbrack1,L\rbrack} {\mathcal{B}}_{\ell,\ell'}^\ast\,,
\end{align}
 where $\nabla_\btheta^2 \Loss(\soln{1}\lbrack\ell,\ell'\rbrack)$ denotes the $(\ell, \ell')^{\text{th}}$ cross diagonal block of the Hessian and where $L$ denotes the network depth.   So for an arbitrary set of layer indices $\mathcal{P}$, which we will use to analyze the cumulative barriers considered in~\citep{adilova2023layer}, we have the expression:
 \begin{align}\label{eq:layerwisebarrier-mostgen}
    {\mathcal{B}}_{\mathcal{P}}^\ast \approx \frac{1}{8} \sum_{\ell,\ell' \,\in\,\mathcal{P}} \, \Delta{\btheta^\ast\lbrack\ell\rbrack}^\top \, \cdot \,\nabla_\btheta^2 \Loss(\soln{1}\lbrack\ell,\ell'\rbrack)\,\cdot \, \Delta\btheta^\ast\lbrack\ell'\rbrack\,,
\end{align}

\clearpage
\section{Additional Results and Details}\label{app:results}
\subsection{Hyperparameter Setup}\label{app:hyperparam}
Unless stated otherwise, following~\citet{frankle2020linear}, we consider a ResNet20 trained on CIFAR10 with batch normalization enabled for $160$ epochs with SGD. The other hyperparameters that were used are a learning rate $0.1$ which is decreased by a factor of $10$ at epochs $80$ and $120$. Besides, other hyperparameters are weight decay  $0.0001$, batch size $128$, momentum $0.9$. 

\subsection{Topography of LMC}
\subsubsection{1D final barrier view}

\begin{figure}[ht!]
    \centering
    \subfigure[The 3 kinds of barriers]{
\includegraphics[width=0.35\textwidth]{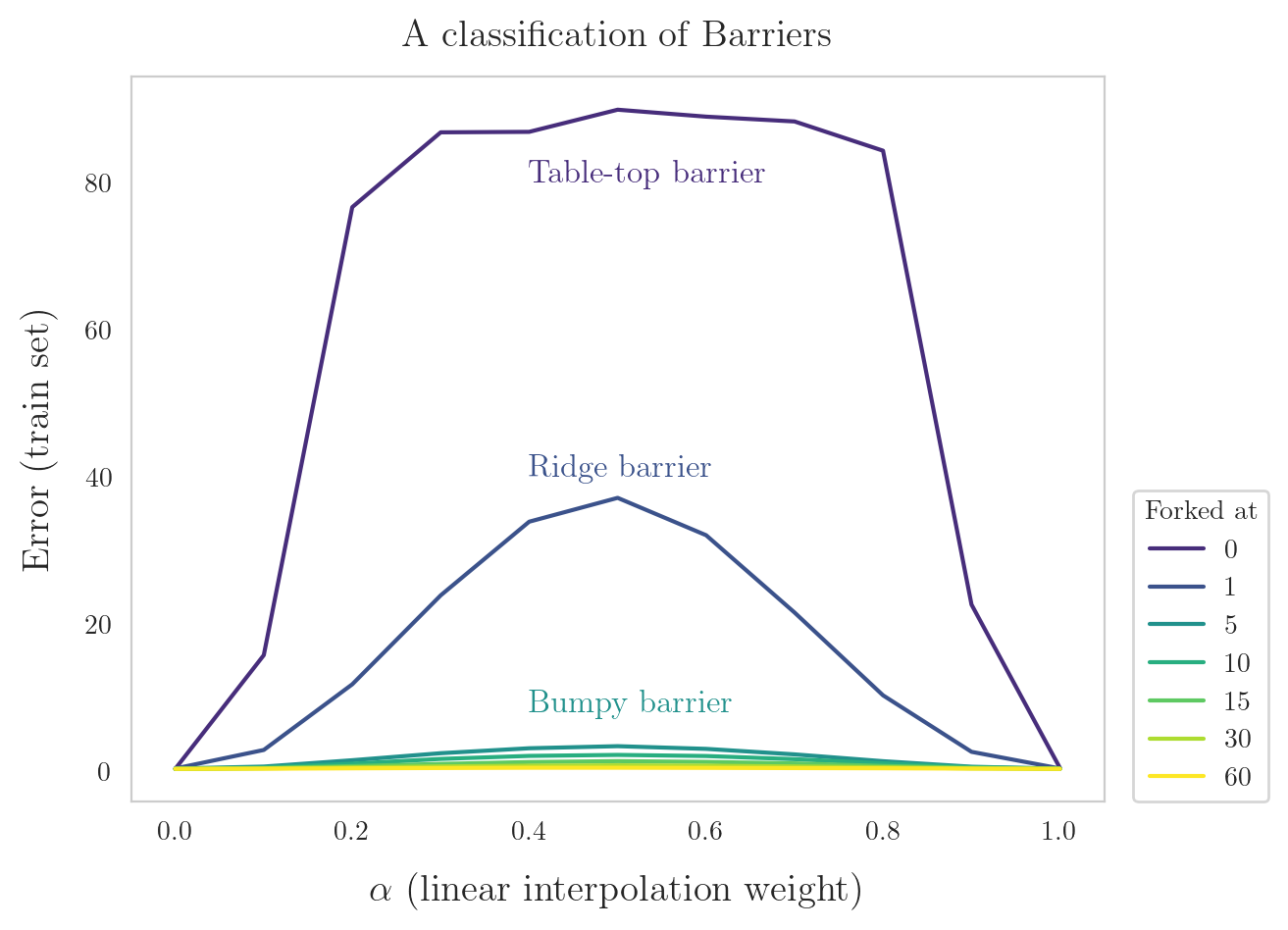}}
  \subfigure[Bumpy barriers when zoomed in]{
    \includegraphics[width=0.35\textwidth]{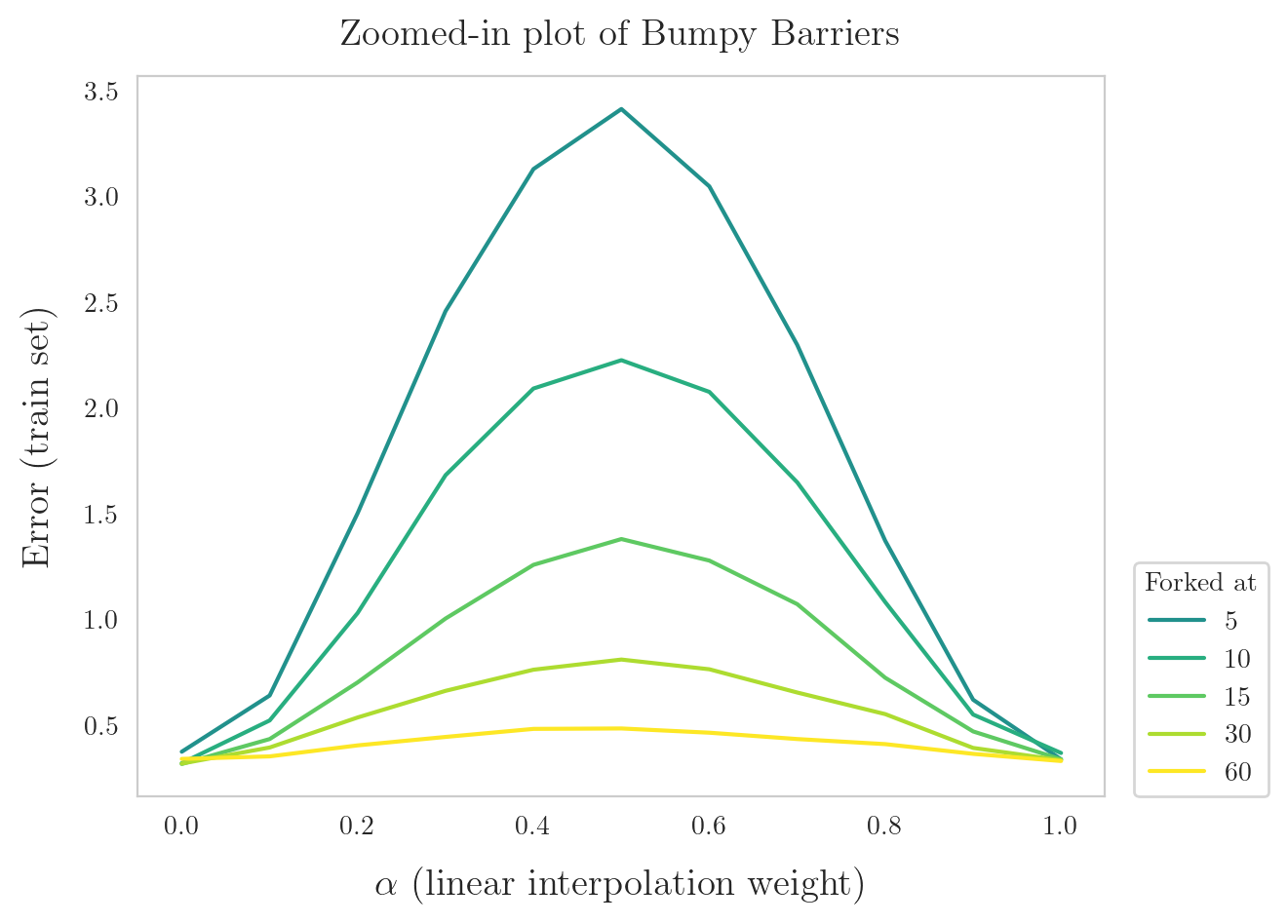}}
    \caption{A one-dimensional summary of the barrier types and their classification.}
    \label{fig:barrier-types}
\end{figure}

\clearpage

\subsubsection{Training Error evolution for other settings}\label{app:train-err-evols}
\begin{figure}[!h]
    \centering
    \includegraphics[width=0.8\textwidth]{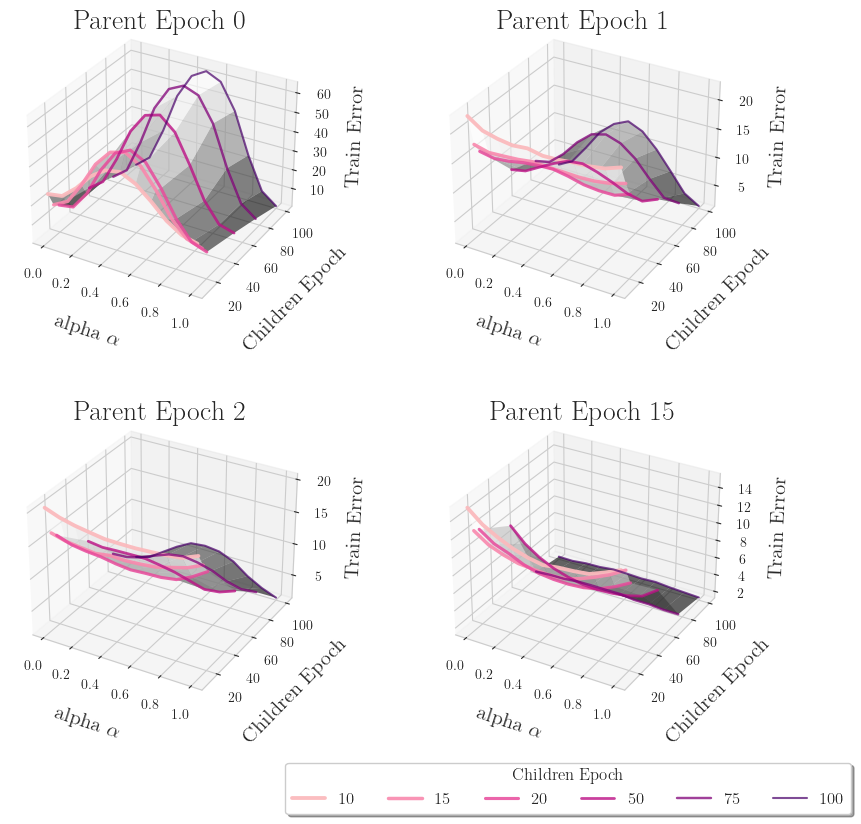}
    \caption{LR $\eta=0.01$, Weight decay $\lambda=0.0001$: The evolution of train error curves, with train error, in 3d when forked at different points.}
    \label{fig:lmc_lowLR_train}
\end{figure}
\begin{figure}[!h]
    \centering
    \includegraphics[width=0.8\textwidth]{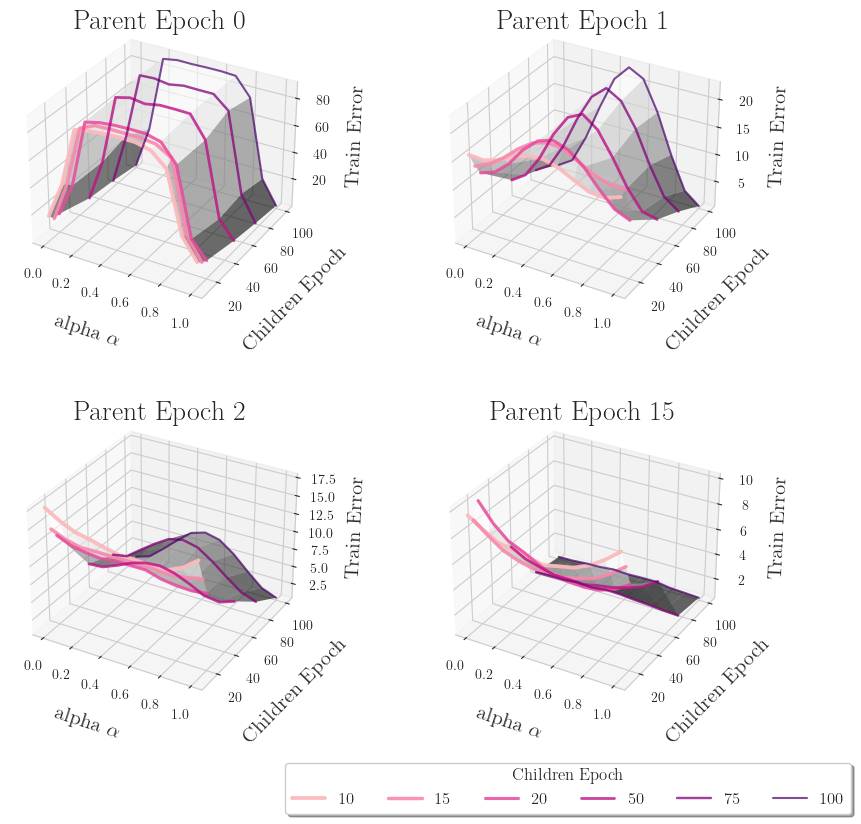}
    \caption{Weight decay $\lambda=0$, LR $\eta=0.1$: The evolution of train error curves, with train error, in 3d when forked at different points.}
    \label{fig:lmc_noWDtrain}
\end{figure}
\clearpage
\subsubsection{Barrier curves evolution}

\begin{figure}[!h]
    \centering
    \includegraphics[width=0.8\textwidth]{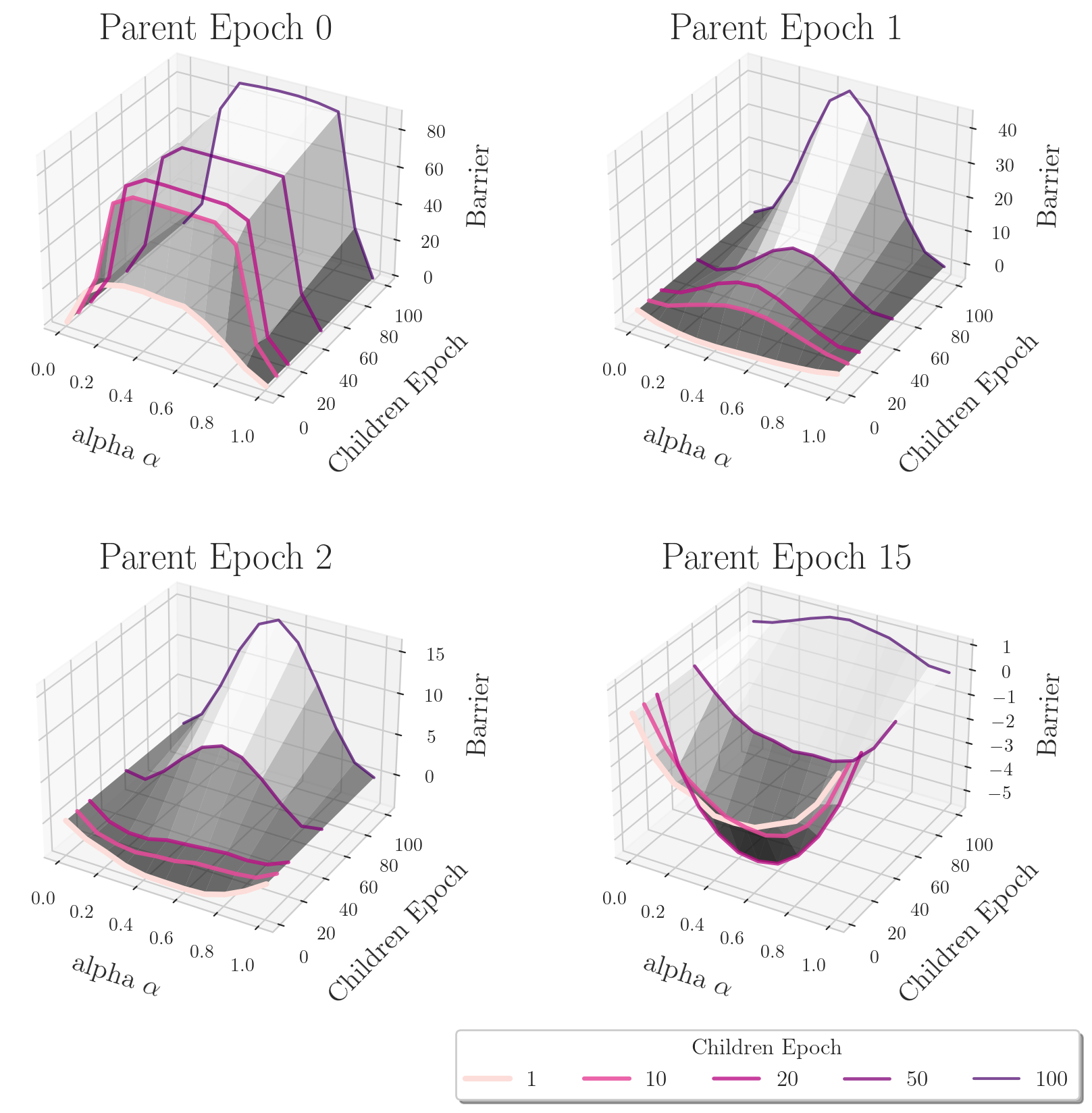}
    \caption{Weight decay $\lambda=0.0001$, LR $\eta=0.1$: The evolution of LMC barrier curves, with train error, in 3d when forked at different points.}
    \label{fig:3d-barrier}
\end{figure}

\begin{figure}[!h]
    \centering
    \includegraphics[width=0.8\textwidth]{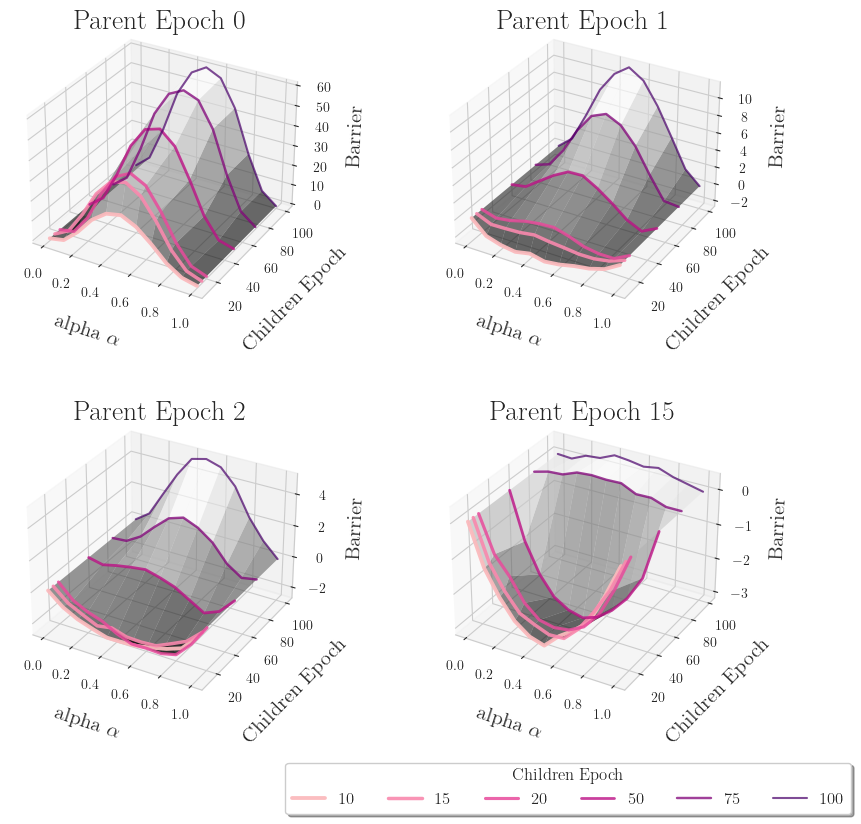}
    \caption{Weight decay $\lambda=0.0001$, LR $\eta=0.01$: The evolution of LMC barrier curves, with train error, in 3d when forked at different points.}
    \label{fig:3d-barrier_lowLR}
\end{figure}
\begin{figure}[!h]
    \centering
    \includegraphics[width=0.8\textwidth]{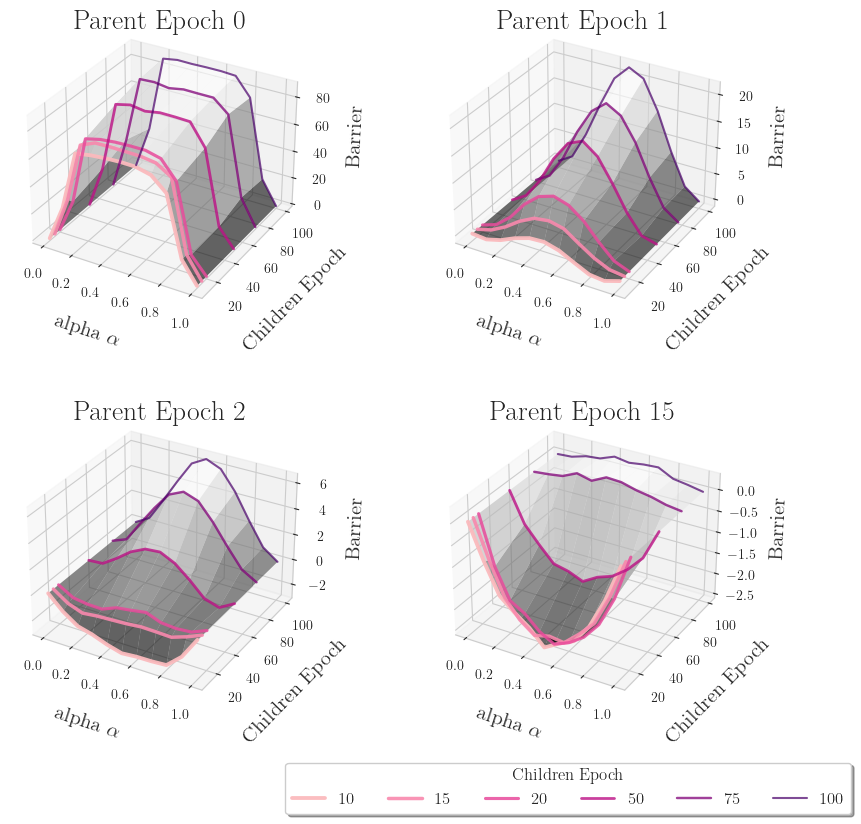}
    \caption{Weight decay $\lambda=0$, LR $\eta=0.1$: The evolution of LMC barrier curves, with train error, in 3d when forked at different points.}
    \label{fig:3d-barrier_noWD}
\end{figure}

\clearpage
\subsubsection{Sibling Geometry Analysis}

\begin{figure}[ht!]
    \centering
\subfigure[Sibling angles, $\eta=0.01$]{
\includegraphics[width=0.35\textwidth]{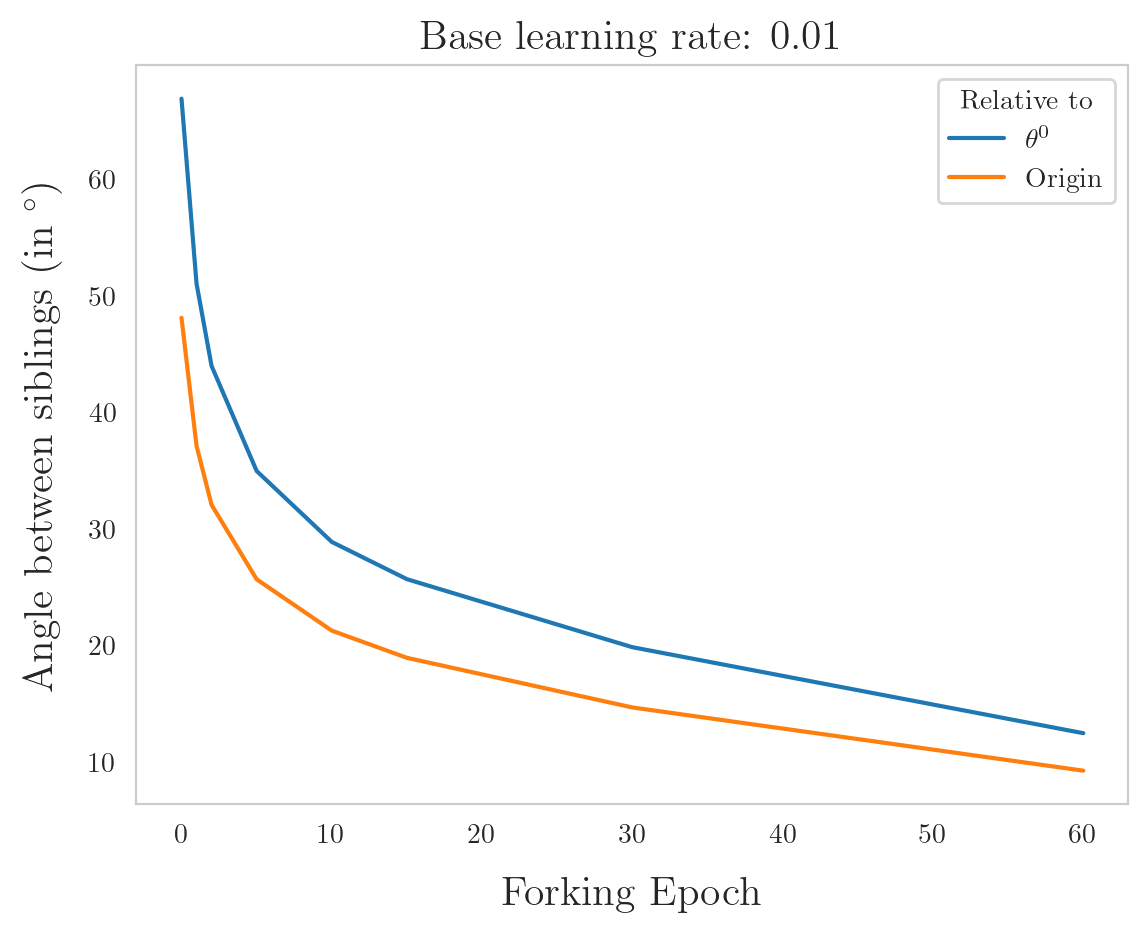}}
 \subfigure[Sibling solution planes, $\eta=0.01$]{
\includegraphics[width=0.36\textwidth]{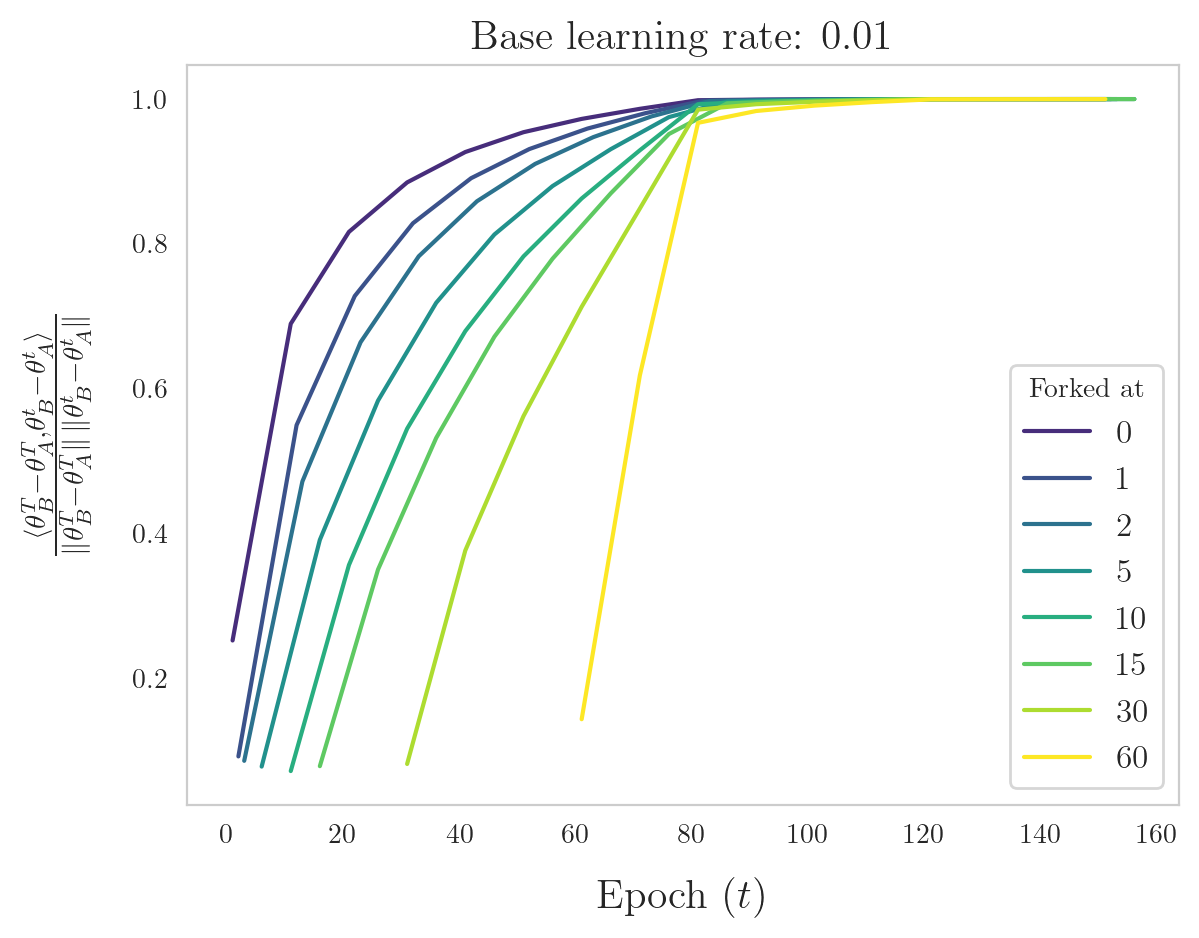}}
 \caption{The angle between sibling solutions (in degrees) as well as the determination of sibling solution planes for different forks, for a small learning rate scenario ($\eta=0.01$).}
 \label{fig:sibling-low-lr}
\end{figure}

\subsection{Barrier Prediction Analysis}
\begin{figure}[ht!]
    \centering
    \subfigure[Predicted vs Actual Barriers]{
	\includegraphics[width=0.4\textwidth]{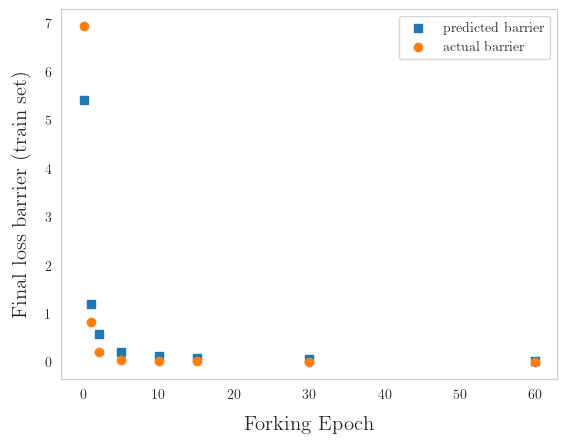}}
 \subfigure[Sibling Distances]{
    	\includegraphics[width=0.46\textwidth]{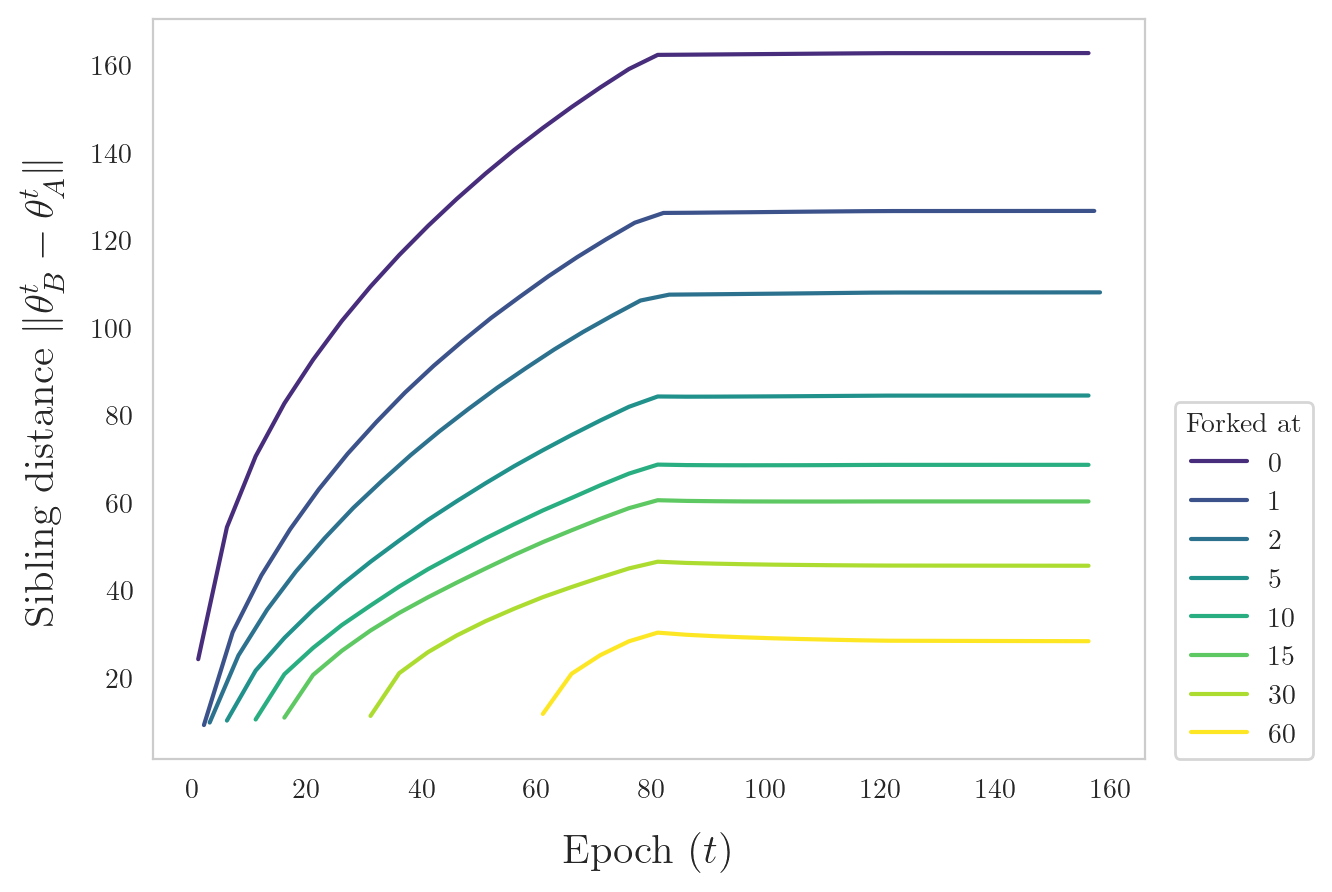}
 }
    \caption{\textbf{Without weight decay. } \textit{Left:} Barrier predictions versus actual barriers over different forking epochs. \textit{Right: }The evolution of the distance between spawned models is plotted, and where  the different lines denote different forking points.%
    }
    \label{fig:forks-dists1-wd}
\end{figure}

\subsection{Layer-wise LMC results}\label{app:layerwise}
\subsubsection{Experimental details for experiments in the main text}\label{app:layerwise-details}

The presented results consider ResNet18 on CIFAR10, without batch normalization layers, with the training hyperparameters being: learning rate $0.05$ and batch size $64$. The full Hessian is computed over the entire training set. While in this experiment Batch Normalization was disabled, in the experiments below, we also consider Batch Normalization. 

\subsubsection{Additional results}\label{app:layerwise-more}
For these experiments, we use the hyperparameter detailed in Section~\ref{app:hyperparam}.
\begin{figure}[ht!]
    \centering
    \subfigure[layer-wise Loss barrier]{
		\includegraphics[width=0.32\textwidth]{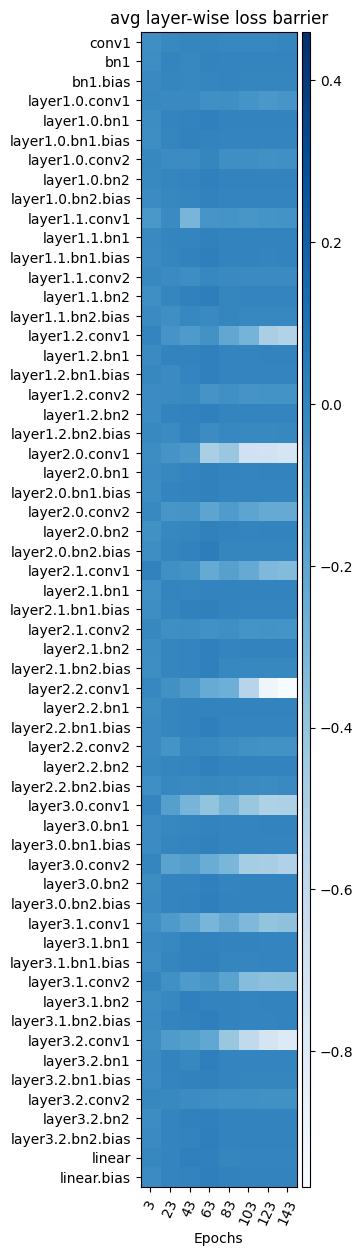}}
  \subfigure[layer-wise Hessian energy]{
    \includegraphics[width=0.32\textwidth]{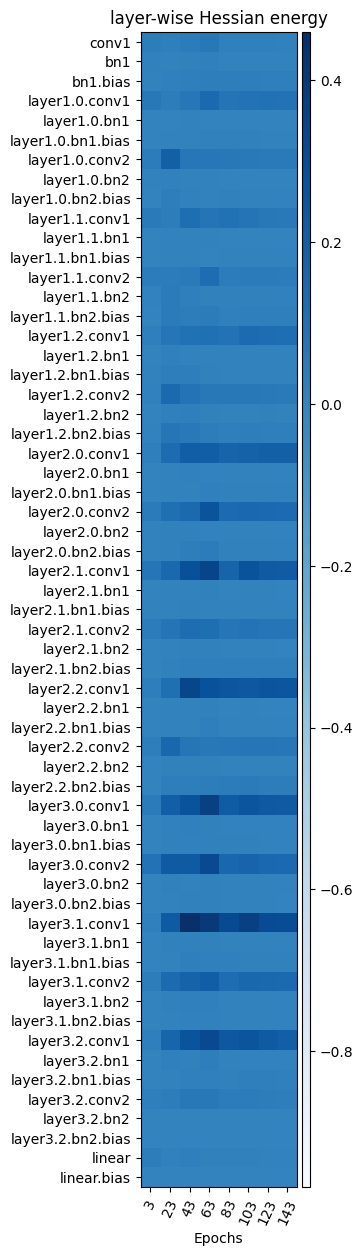}}
  \subfigure[deltas]{
    \includegraphics[width=0.315\textwidth]{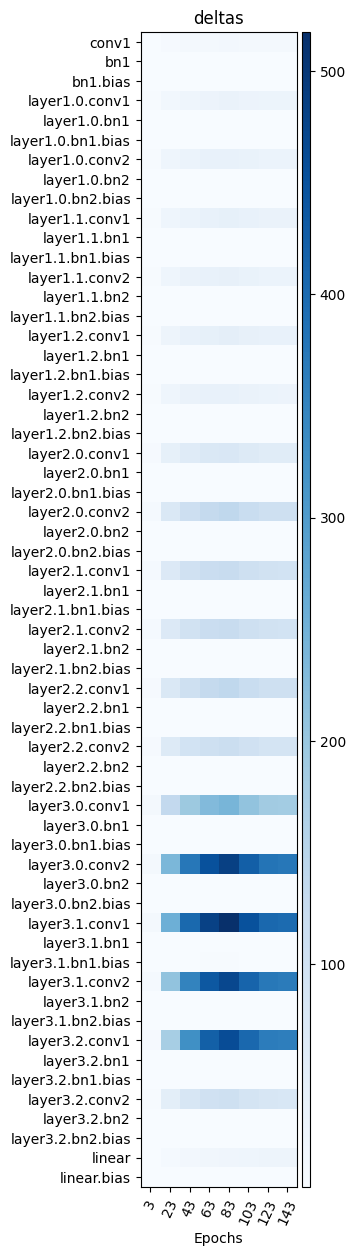}}
    \caption{\textbf{(Left)} The layer-wise loss barrier as per~\cite{adilova2023layer}. \textbf{(Middle):} Layer-wise Hessian Energy, computed as per Proposition~\ref{theorem:llmc-barrier} during the course of training. Forked on epoch 2. \textbf{(Right):} Distance between layers.}
\end{figure}

\begin{figure}[ht!]
    \centering
    \subfigure[layer-wise Loss barrier]{
		\includegraphics[width=0.32\textwidth]{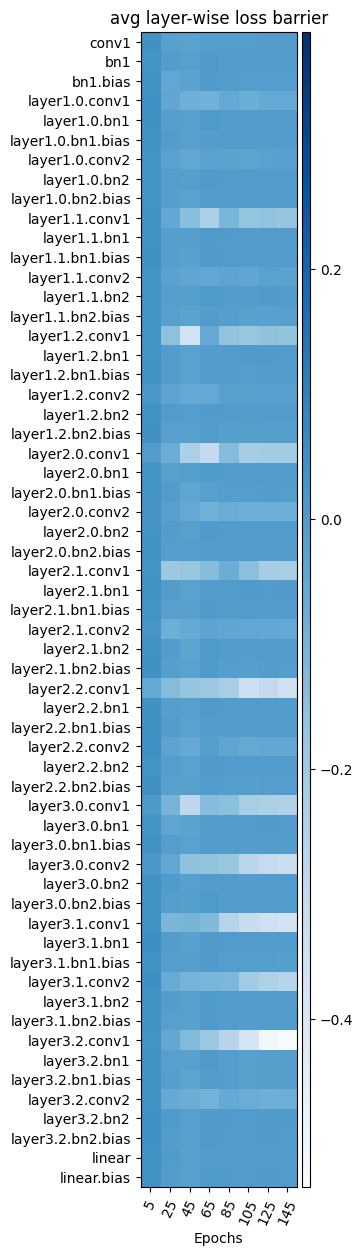}}
  \subfigure[layer-wise Hessian energy]{
    \includegraphics[width=0.32\textwidth]{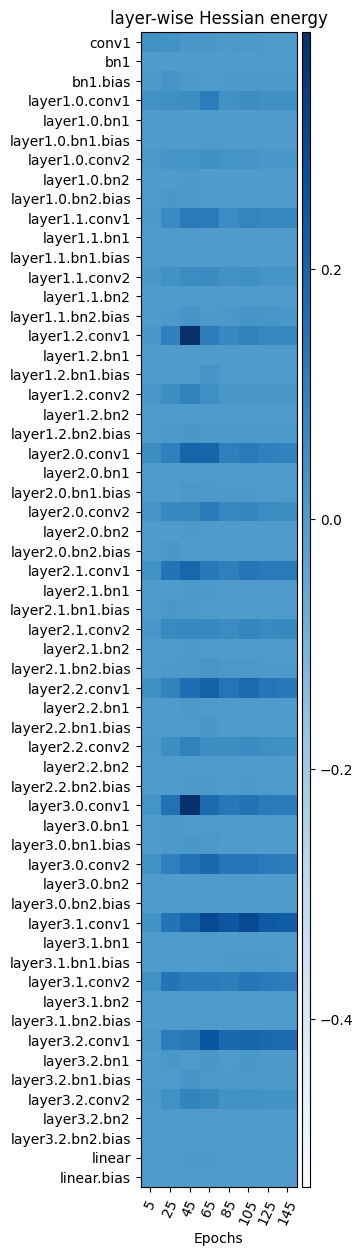}}
  \subfigure[deltas]{
    \includegraphics[width=0.315\textwidth]{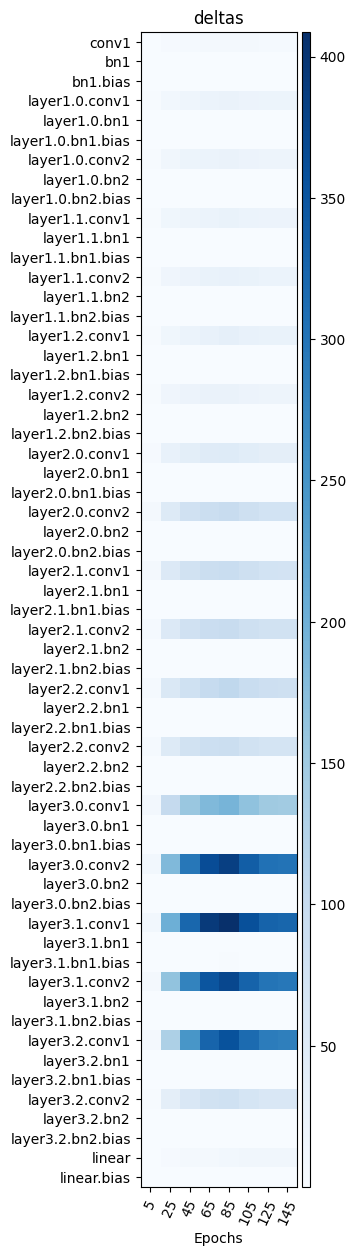}}
    \caption{\textbf{(Left)} The layer-wise loss barrier as per~\cite{adilova2023layer}. \textbf{(Middle):} Layer-wise Hessian Energy, computed as per Proposition~\ref{theorem:llmc-barrier} during the course of training. Forked on epoch 4. \textbf{(Right):} Distance between layers.}
\end{figure}

\begin{figure}[ht!]
    \centering
    \subfigure[layer-wise Loss barrier]{
		\includegraphics[width=0.32\textwidth]{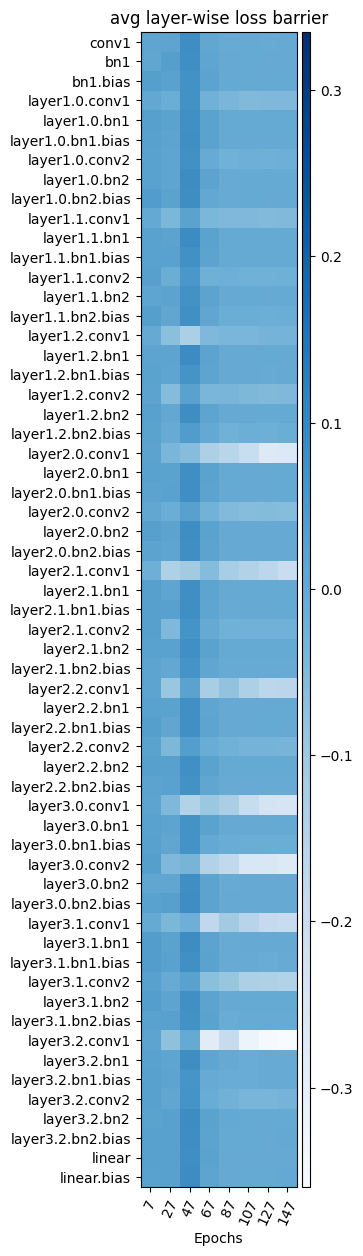}}
  \subfigure[layer-wise Hessian energy]{
    \includegraphics[width=0.32\textwidth]{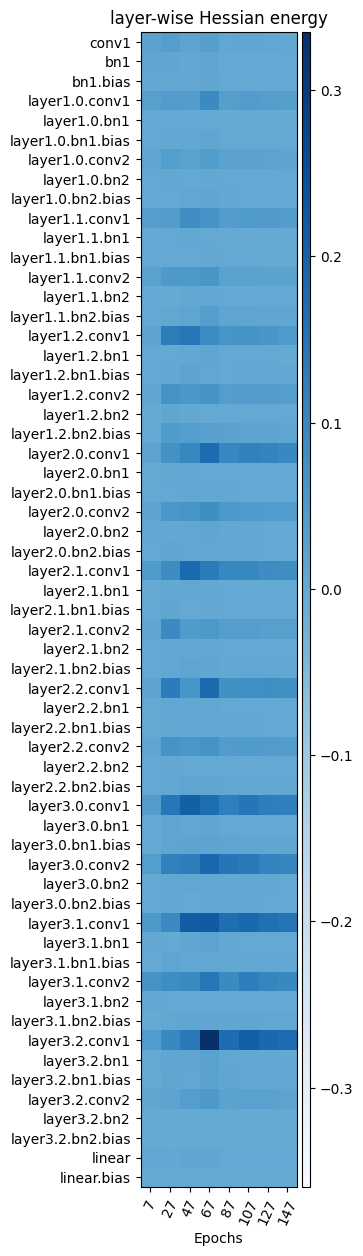}}
  \subfigure[deltas]{
    \includegraphics[width=0.315\textwidth]{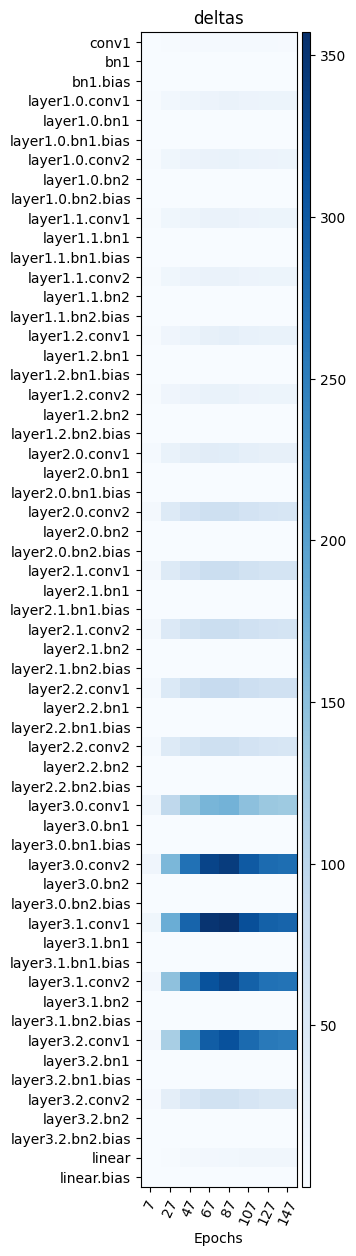}}
    \caption{\textbf{(Left)} The layer-wise loss barrier as per~\cite{adilova2023layer}. \textbf{(Middle):} Layer-wise Hessian Energy, computed as per Proposition~\ref{theorem:llmc-barrier} during the course of training. Forked on epoch 6. \textbf{(Right):} Distance between layers.}
\end{figure}

\begin{figure}[ht!]
    \centering
    \subfigure[layer-wise Loss barrier]{
		\includegraphics[width=0.32\textwidth]{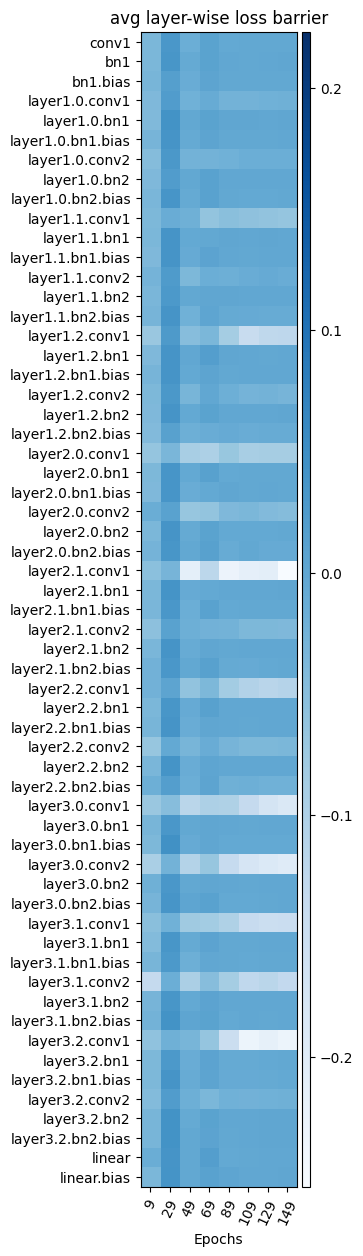}}
  \subfigure[layer-wise Hessian energy]{
    \includegraphics[width=0.32\textwidth]{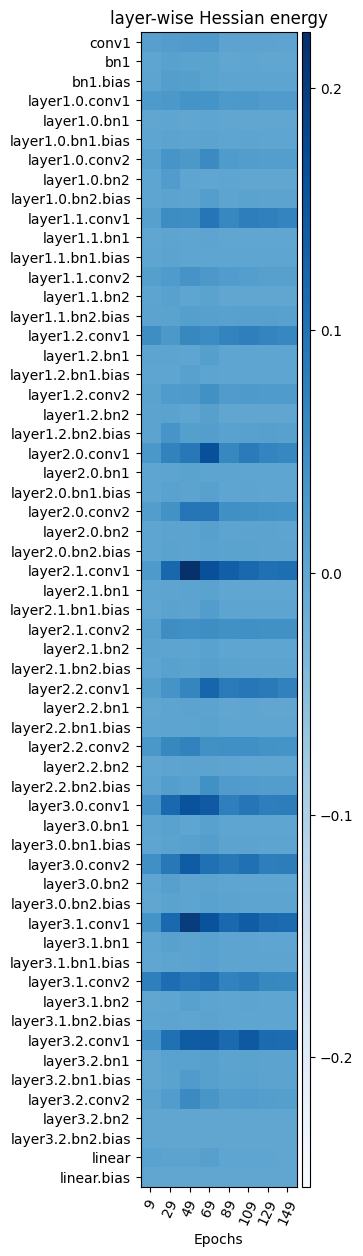}}
  \subfigure[deltas]{
    \includegraphics[width=0.315\textwidth]{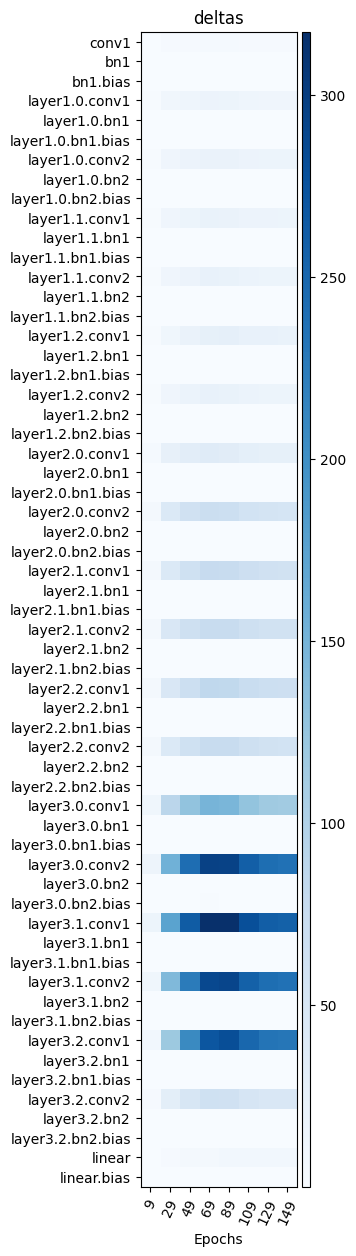}}
    \caption{\textbf{(Left)} The layer-wise loss barrier as per~\cite{adilova2023layer}. \textbf{(Middle):} Layer-wise Hessian Energy, computed as per Proposition~\ref{theorem:llmc-barrier} during the course of training. Forked on epoch 8. \textbf{(Right):} Distance between layers.}
\end{figure}

\begin{figure}[ht!]
    \centering
    \subfigure[layer-wise Loss barrier]{
		\includegraphics[width=0.32\textwidth]{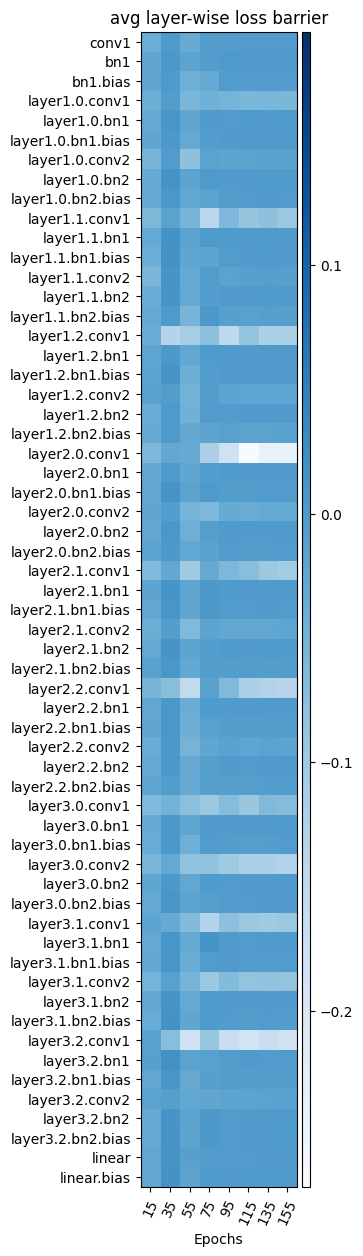}}
  \subfigure[layer-wise Hessian energy]{
    \includegraphics[width=0.32\textwidth]{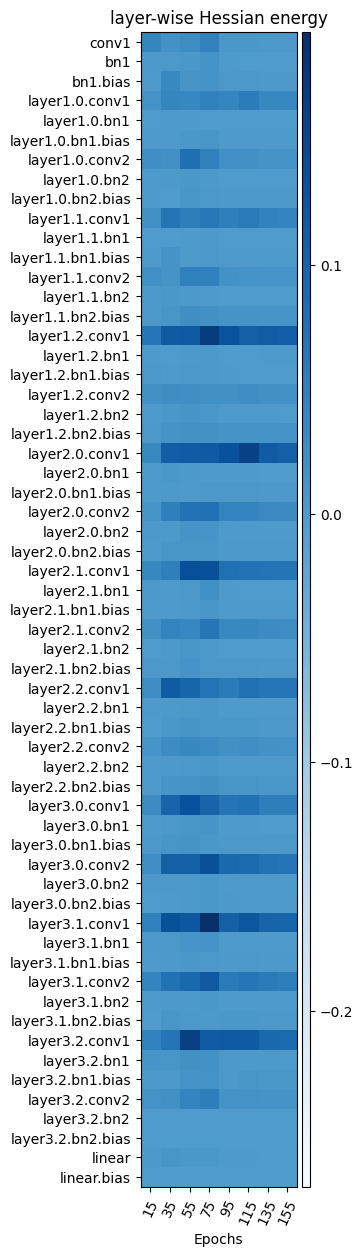}}
  \subfigure[deltas]{
    \includegraphics[width=0.315\textwidth]{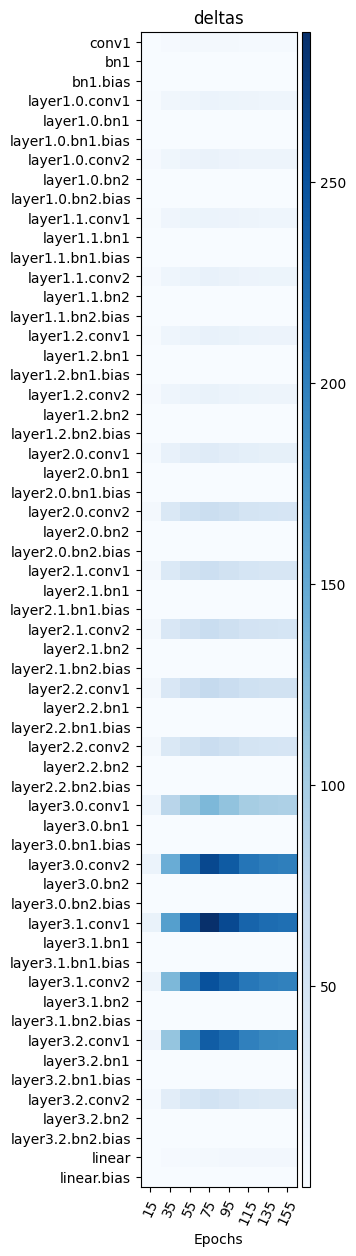}}
    \caption{\textbf{(Left)} The layer-wise loss barrier as per~\cite{adilova2023layer}. \textbf{(Middle):} Layer-wise Hessian Energy, computed as per Proposition~\ref{theorem:llmc-barrier} during the course of training. Forked on epoch 14. \textbf{(Right):} Distance between layers.}
\end{figure}

\begin{figure}[ht!]
    \centering
    \subfigure[layer-wise Loss barrier]{
		\includegraphics[width=0.32\textwidth]{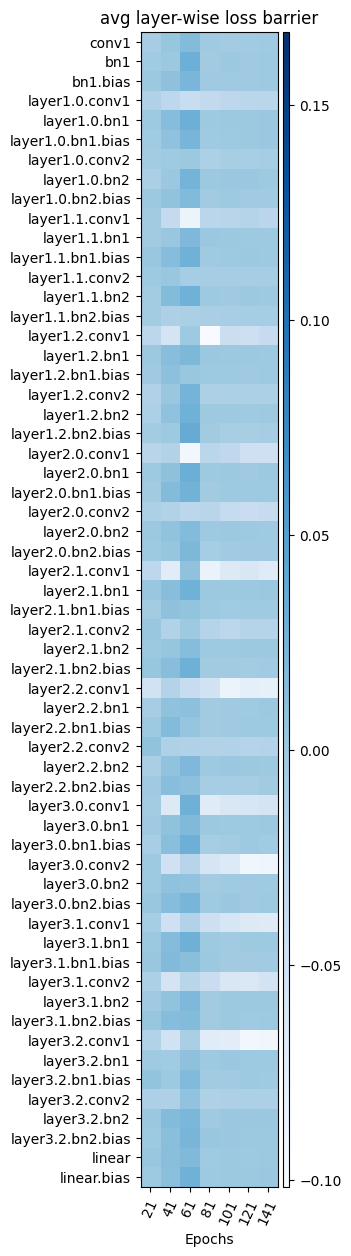}}
  \subfigure[layer-wise Hessian energy]{
    \includegraphics[width=0.32\textwidth]{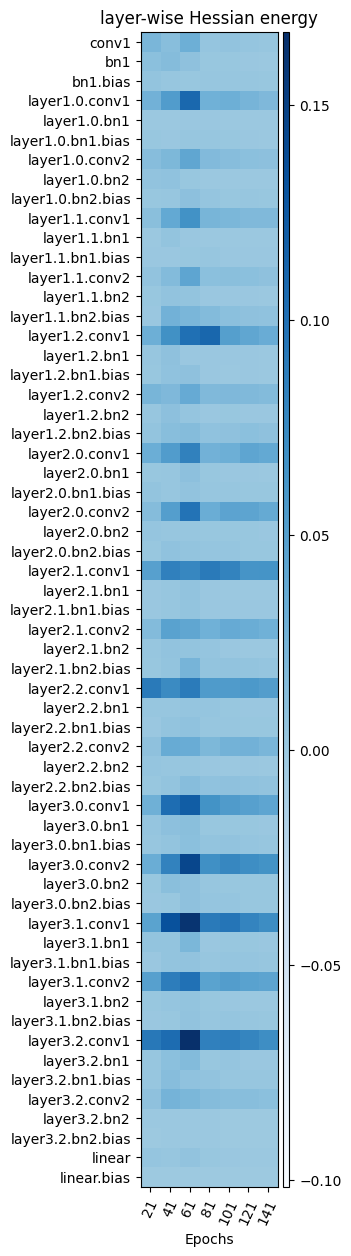}}
  \subfigure[deltas]{
    \includegraphics[width=0.305\textwidth]{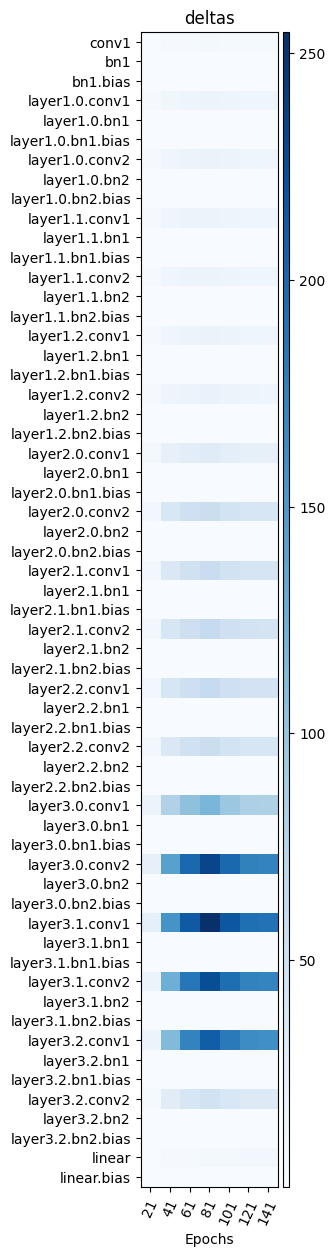}}
    \caption{\textbf{(Left)} The layer-wise loss barrier as per~\cite{adilova2023layer}. \textbf{(Middle):} Layer-wise Hessian Energy, computed as per Proposition~\ref{theorem:llmc-barrier} during the course of training. Forked on epoch 20. \textbf{(Right):} Distance between layers.}
\end{figure}

\end{document}